\documentclass{article} 
\usepackage{iclr2027_arxiv,times}

\usepackage{amsmath,amsfonts,bm}

\newcommand{\figleft}{{\em (Left)}}

\newcommand{\figright}{{\em (Right)}}

\def\eqref#1{equation~\ref{#1}}

\def\1{\bm{1}}

\DeclareMathAlphabet{\mathsfit}{\encodingdefault}{\sfdefault}{m}{sl}
\SetMathAlphabet{\mathsfit}{bold}{\encodingdefault}{\sfdefault}{bx}{n}

\usepackage{hyperref}
\usepackage{url}

\usepackage{hyperref}
\usepackage{url}
\usepackage{booktabs}
\usepackage{graphicx}
\usepackage{color, colortbl}
\usepackage{subcaption}
\usepackage{multirow}
\usepackage{xspace}
\usepackage{courier}
\usepackage{listings}
\usepackage{algorithm}

\title{The Devil is in the Spectrum Bias: Spectrum-Balanced Feature Matching for Robust Representation Distillation}

\author{
    Kuniaki Saito\thanks{Corresponding author.}, 
    Yoshitaka Ushiku
    \\
    OMRON SINIC X Corporation
}

\newcommand{\ie}{i.e.,\xspace}

\newcommand{\ours}{SpecMatch\xspace}
\definecolor{ourscolor}{RGB}{235,243,250}
\iclrfinalcopy 
\begin{document}

\maketitle

\begin{abstract}
Large visual foundation models have demonstrated remarkable transferability across a wide range of downstream tasks. To deploy such models efficiently, feature matching has become a popular knowledge distillation approach that transfers teacher representations to smaller student models without requiring labeled data. However, we show that the conventional feature matching objective with L2-distance is inherently biased toward reconstructing dominant spectral directions of the teacher representation, while under-optimizing low-variance directions that often contain task-relevant information. To address this, we propose Spectrum-Balanced Feature Matching, \ours, a simple objective that adaptively emphasizes under-optimized spectral directions while preserving the relative importance of dominant directions. \ours is easy to implement and introduces negligible computational overhead. Extensive experiments on image recognition demonstrate that \ours consistently improves downstream adaptation across diverse tasks, including image classification, anomaly detection, medical image analysis, and domain generalization. In particular, \ours outperforms conventional feature matching in 40 of 42 teacher--student and training-setting combinations, while consistently improving over the original student model in all settings. We further demonstrate that the proposed objective generalizes beyond vision, improving downstream performance across six protein understanding tasks.
\end{abstract}

\section{Introduction}
Recent advances in large-scale representation learning have led to the emergence of large foundation models trained on billions of images~\citep{oquab2023dinov2,simeoni2025dinov3,bolya2026perception,chuang2026meta,xu2024demystifying,radford2021learning}. Such models have demonstrated remarkable transferability across a wide range of downstream tasks, including object recognition, medical image analysis, and domain generalization. As foundation models continue to grow in scale and capability, they increasingly serve as a source of visual knowledge for downstream applications.

\begin{figure}[t]
    \centering   
    \includegraphics[height=0.3\linewidth]{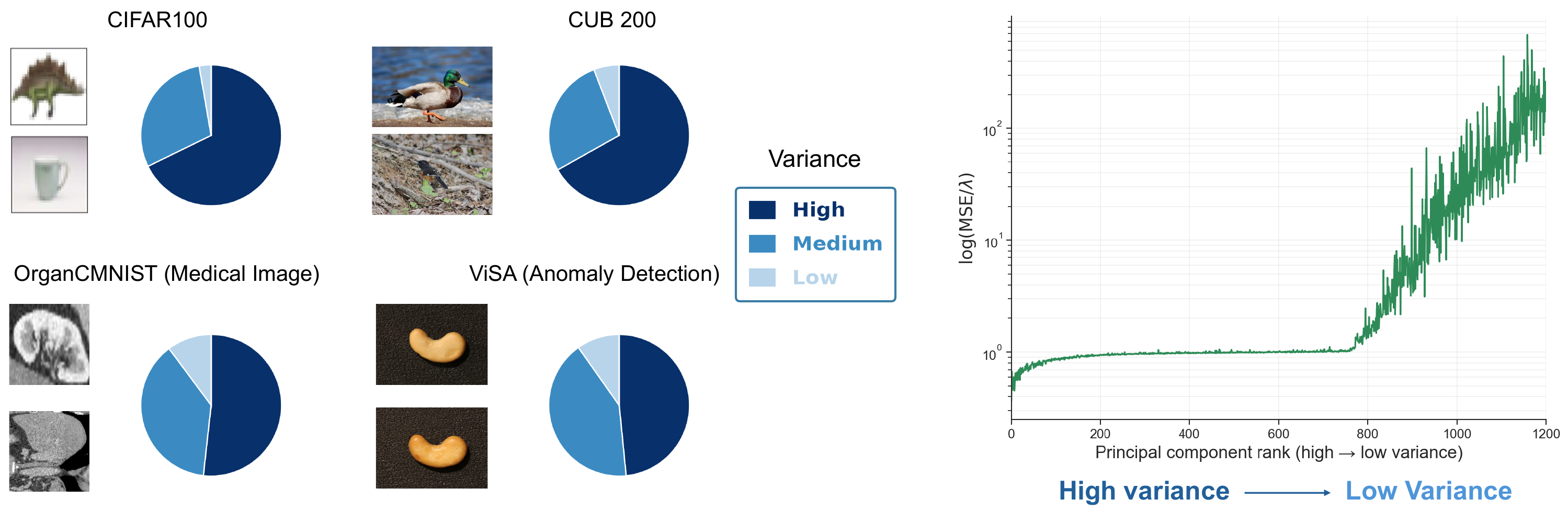}
        \label{fig:task_motivation}
    \vspace{-1mm}
    \caption{\textbf{\figleft:} Contribution of principal component (PC) groups to downstream tasks. PCs are sorted by explained variance and divided into three equal-sized groups. Different tasks rely on different spectral regions, motivating the preservation of a broad representation spectrum during distillation. \textbf{\figright:} Variance-normalized reconstruction error of vanilla L2 feature matching. Reconstruction deteriorates toward higher-rank, lower-variance PCs, revealing a bias toward high-variance components that may discard information useful for downstream tasks. We further provide evidence in Fig.~\ref{fig:pca_percent_accuracy} that lower-variance PCs do not merely represent noise but can substantially improve performance on certain downstream tasks.}
    \label{fig:motivation}
\end{figure}

However, directly deploying these large models is often impractical due to computational constraints. Consequently, an important challenge is how to effectively transfer the rich knowledge in large models to smaller and more efficient models. Knowledge distillation has emerged as a promising solution to this problem, enabling student models to inherit useful representations from powerful teachers~\citep{hinton2015distilling,jang2025vl2lite,lee2025customkd,ranzinger2024radio,zhang2025accessing}. 
Feature matching is a popular distillation approach since it transfers teacher representations to a student model and can be applied to diverse scenarios in a label-free manner~\citep{sariyildiz2024unic,chen2021distilling,wang2019distilling}. However, standard feature matching objectives treat the teacher representation as a whole, without explicitly accounting for how information is distributed across different feature directions. This raises an important question: \textit{does feature matching effectively transfer all information encoded in the teacher representation, including information that may be important for downstream tasks?} Our preliminary analysis reveals an important tension: teacher representations learned from large-scale pre-training are highly anisotropic, with a small number of principal directions dominating the feature variance; however, non-dominant directions can also contain information that contributes substantially to downstream tasks. As illustrated in Fig.~\ref{fig:motivation}, different downstream tasks rely on different spectral regions of the teacher representation, and important task-relevant information can reside in non-dominant directions.

Motivated by this observation, we identify a limitation of conventional feature matching in transferring teacher knowledge to a student. We theoretically and empirically show that feature matching with L2 distance is biased toward reproducing dominant spectral directions: its learning signals scale with teacher feature variance, leading student models to prioritize high-variance directions over low-variance ones. As a result, the student may fail to preserve information that is useful for downstream transfer. Indeed, we empirically observe settings in which vanilla feature matching fails to improve downstream performance. To address this issue, we propose Spectrum-Balanced Feature Matching, \textit{\ours}, a simple objective that adaptively emphasizes under-optimized spectral directions while preserving the relative importance of dominant directions. \ours is easy to implement, introduces negligible computational overhead, and can be incorporated into existing feature matching frameworks with minimal modification.


Extensive experiments on image recognition demonstrate that \ours generally outperforms conventional feature-matching baselines in a downstream adaptation setting, where distillation is performed separately for each target dataset using only its unlabeled images. Across diverse tasks—including image classification, anomaly detection, medical image classification, and domain generalization—our method outperforms vanilla feature matching in 40 out of 42 teacher–student and training-setting combinations, while improving over the original student model in all 42 settings. The method remains effective on large-scale datasets such as ImageNet, as well as on challenging datasets such as iNaturalist, which contains a large number of categories and a highly imbalanced class distribution. Furthermore, the proposed objective generalizes to protein foundation models, achieving consistent improvements across six downstream protein understanding tasks.
\vspace{-3mm}
\section{Related Work}
\vspace{-3mm}
\noindent\textbf{Knowledge distillation} is a popular approach to transfer knowledge between different models by training one model to mimic the outputs of another model. Various approaches have been proposed and some employ task logits~\citep{hinton2015distilling}, others utilize intermediate features or embeddings~\citep{heo2019comprehensive, tian2020contrastive}, relations between samples~\citep{park2019relational}. Some of the self-supervised learning approaches include learning mechanisms similar to model distillation~\citep{oquab2023dinov2,simeoni2025dinov3}. Feature distillation approaches are common in distillation literature while they do not account for transferring rich teacher representations~\citep{sariyildiz2024unic, heo2019comprehensive,chen2021distilling,romero2015fitnets}. We mainly focus on how to adapt the student model to a downstream task by leveraging a huge teacher model, using unlabeled data from the downstream domain. The downstream tasks can require diverse features for the performance boost, thus transferring diverse features is desirable. Our setting is close to \citep{vemulapalli2024knowledge,jang2025vl2lite} in that they assume a label-scarce scenario for cheaper adaptation. While \cite{vemulapalli2024knowledge} focus on obtaining the auxiliary dataset that helps the adaptation to the target domain, our focus is on developing an objective that achieves rich feature transfer. \ours is simple, yet effective in improving the performance of the student model by allowing students to learn rich features. 

\noindent\textbf{Normalization in knowledge distillation.} 
\cite{miles2024v,miles2024understanding,lee2018self} employ feature normalization or whitening in knowledge distillation. Our theoretical analysis provides an intuition for why whitening can facilitate feature matching: by equalizing the variance of the teacher representation across spectral directions, it prevents a small number of high-variance directions from dominating the objective. However, our analysis shows that whitening does not consistently improve downstream transfer and can even cause substantial degradation on several datasets. These results suggest that completely equalizing spectral directions can be overly aggressive, since dominant directions may remain important for downstream tasks. In contrast, \ours is designed to reduce the excessive optimization focus on high-variance directions without enforcing equal contributions across the spectrum. Specifically, it moderately emphasizes low-variance directions without fully equalizing their contributions with those of high-variance directions.

\noindent\textbf{Richness of feature representations.}
The richness of learned representations has been extensively studied in the self-supervised learning (SSL) literature. RankMe~\citep{garrido2023rankme} showed that the effective rank of learned representations is strongly correlated with downstream performance, suggesting that high-rank representations capture richer semantic information. Motivated by this observation, several SSL methods have been proposed to prevent rank collapse or maintain a high effective rank of learned representations~\citep{zbontar2021barlow,jing2021understanding,he2022exploring}. Our work shares a similar motivation in that we aim to preserve the richness of teacher representations during knowledge transfer. However, rather than designing a self-supervised objective to improve the representation itself, we develop a distillation objective that transfers the rich representations of a powerful teacher model to a student model. 

\vspace{-5mm}
\section{Revisiting Feature Matching for Representation Transfer}
\vspace{-2mm}
\subsection{Task-Relevant Information is Distributed Across the Spectrum}
We analyze the contribution of each principal component to downstream classification. Let $z\in\mathbb{R}^{d}$ denote the teacher representation, $U=[u_1,\ldots,u_d]$ the orthonormal PCA basis computed from the teacher representations (ordered by decreasing variance), and $W=[w_1,\ldots,w_C]^\top$ the weight matrix of a linear classifier trained for a downstream task, where $C$ is the number of classes. Since $U$ is orthonormal, each classifier weight can be decomposed as $w_c=\sum_{k=1}^{d}(u_k^\top w_c)u_k$. We quantify the contribution of the $k$-th principal component by measuring the energy of the classifier weights projected onto that direction: $I_k=\sum_{c=1}^{C}(u_k^\top w_c)^2.$ Fig.~\ref{fig:motivation} (left) partitions the principal
components into consecutive rank intervals and compute the normalized contribution within each interval: $C_m
=
\frac{
\sum_{k\in\mathcal{B}_m} I_k
}{
\sum_{k=1}^{d}I_k
}$,
where $\mathcal{B}_m$ denotes the $m$-th rank interval. The darker shades of blue indicate principal component groups with larger teacher variance. For the CIFAR-10, the task-relevant directions are strongly aligned with the dominant principal components of the teacher representation, while higher-rank components contribute little to the classifier. In contrast, for anomaly detection and medical image classification, which require recognizing subtle visual patterns, lower-variance principal components also exhibit substantial contributions. These suggest that transferring only the dominant directions of the teacher can be insufficient for downstream tasks. Instead, effective representation transfer should preserve information across a broader spectrum of principal components.

\vspace{-3mm}
\subsection{Insights into Vanilla Feature Matching}\label{sec:motivation_insight}
\vspace{-2mm}
We analyze the feature matching with L2-distance from the perspective of the principal components of the teacher representation. Let
$z_b^t,z_b^s\in\mathbb{R}^d$
denote the teacher and student representations for the $b$-th sample, respectively.
Applying PCA to the teacher features yields an orthonormal basis
$\{u_k\in\mathbb{R}^d\}_{k=1}^{d}$,
where the teacher representation is expressed as $z_b^t=\sum_{k=1}^{d}c_{b,k}^t u_k$,
and $\lambda_k=\mathrm{Var}(c_k^t)$ denotes the variance along the $k$-th principal direction.

Since the PCA basis is orthonormal, feature matching is equivalent to minimizing the reconstruction error in the PCA space:
\begin{equation}
e_k
=
\frac1B
\sum_{b=1}^{B}
(\tilde c_{b,k}^{\,s}-c_{b,k}^{\,t})^2,
\qquad
\mathcal{L}_{\mathrm{FM}}
=
\sum_{k=1}^{d}
e_k
\label{eq:error_axis}
\end{equation}
where $\tilde c_{b,k}^{\,s}$ is the student prediction along the $k$-th principal direction. At the early stage of training, the expected reconstruction error satisfies
\begin{equation}
\mathbb{E}[e_k]
\approx
\lambda_k+C,
\label{eq:error_lambda}
\end{equation}
where $C$ is approximately constant across principal components (See Sec.~\ref{sec:theory} for proof). Combining Eq.~\ref{eq:error_lambda} with the gradient analysis (See Sec.~\ref{sec:theory}), we obtain
\begin{equation}
\mathbb{E}
\!\left[
\left\|
\nabla_{\tilde{\mathbf c}_{:,k}}
\mathcal L_{\mathrm{FM}}
\right\|_2^2
\right]
\propto
\lambda_k+C.
\label{eq:gradient_bias}
\end{equation}
This shows that the gradient norm along each principal direction \(k\) scales with \(\lambda_k\), i.e., the variance of the teacher features along that direction. This demonstrates that the optimization dynamics of feature matching allocate larger optimization signals to principal components with larger teacher variance, which can hinder the reconstructing target features with higher ranks, low-variance principal directions \footnote{Fig.~\ref{fig:eigenvalue_mse} shows that, even after many training iterations, $e_k$ remains approximately proportional to $\lambda_k$ for high-variance directions. These directions continue to dominate the optimization throughout training.}.
In fact, the right of Fig.~\ref{fig:motivation} shows that a model trained with vanilla feature matching struggles to accurately reconstruct the teacher representations along higher-rank directions. 
\vspace{-3mm}
\section{\ours: Spectrum-Balanced Feature Matching}\label{sec:method} 
\vspace{-2mm}
To mitigate the spectral bias explained above, we propose Spectrum-Balanced Feature Matching, \ours. Our key observation is that reconstruction errors tend to be larger along dominant principal directions due to their higher variance. Consequently, the normal feature matching allocates disproportionate learning signals to these directions. To address this, we introduce a simple framework that explicitly mitigates the imbalance in the contributions using their reconstruction errors.

\begin{figure}[t]
    \centering
       \includegraphics[width=\linewidth]{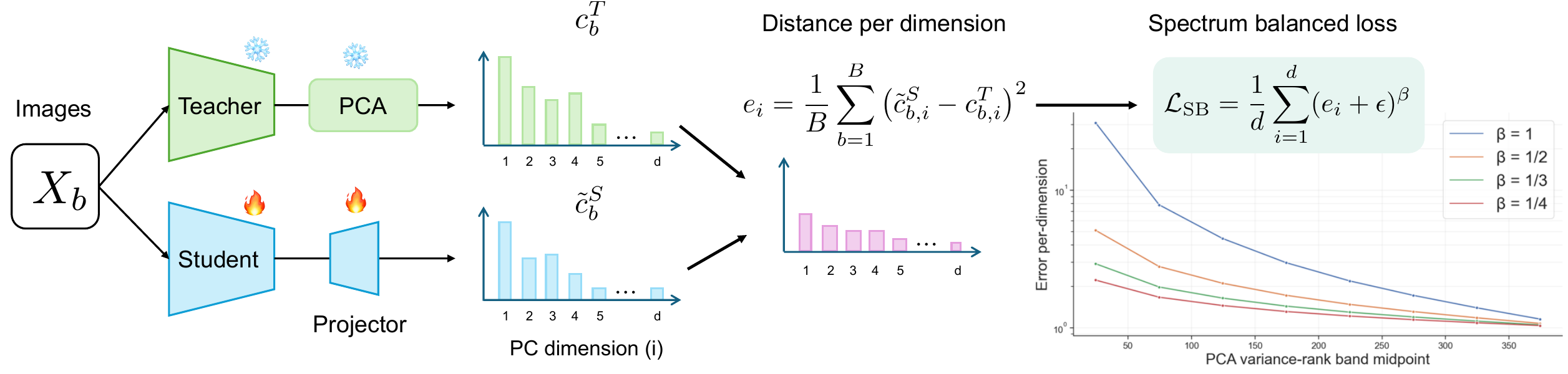}
       \vspace{-7mm}
    \caption{
    \small
    \textbf{\figleft:} Description of the proposed objective. \textbf{\figright:} The illustration of the error (y-axis) on different principal components (x-axis).}
    \label{fig:method}
\end{figure}

\noindent\textbf{Predicting coefficients of the PCA basis.}
As shown in the left of Fig.~\ref{fig:method}, the student predicts the coefficients of the teacher representation in the PCA basis, rather than directly regressing the teacher features. This formulation makes the reconstruction error along each principal direction explicit, allowing us to balance the loss across spectral directions. Both the teacher model and the PCA basis are fixed during training. We employ a linear classifier as the prediction head. 

\noindent\textbf{Proposed Objective.}
To quantify the reconstruction error along each principal direction, we first compute the average teacher–student error \(e_k\) within each mini-batch, as defined in Eq.~\ref{eq:error_axis}. As indicated by Eq.~\ref{eq:error_lambda}, \(e_k\) scales with the variance \(\lambda_k\) of the corresponding teacher direction. Rather than linearly aggregating these per-direction errors, we employ a sub-linear aggregation: \begin{equation} \mathcal{L}_{\mathrm{SB}} = \sum_{i=1}^{d} (e_k+\epsilon)^\beta, \qquad 0<\beta<1, \end{equation} where $\epsilon$ is a small constant for numerical stability. 
This transformation reduces the relative influence of large-error directions and increases that of under-optimized directions. 
$\beta$ is an important hyperparameter that controls the degree of emphasis placed on low-energy components. As \(\beta\) approaches 1, the objective becomes equivalent to the standard L2 feature-matching loss. As \(\beta\) approaches 0, the contributions of low- and high-energy components become more uniform.
Importantly, because \((e_k+\epsilon)^\beta\) remains monotonically increasing in \(e_k\), directions with large variance and reconstruction error still receive greater emphasis; the objective only moderates their dominance to achieve a better balance across the spectrum. The right panel of Fig.~\ref{fig:method} illustrates \((e_k+\epsilon)^\beta\) for different values of \(\beta\) on real data, with principal components ordered from the highest to the lowest teacher variance. Empirically, high-variance components tend to exhibit larger reconstruction errors, whereas low-variance components incur substantially smaller errors. Our objective compresses this disparity, balancing the contributions of different principal components while preserving the greater importance of dominant directions. The pseudo-code is available in Alg.~\ref{alg:specmatch}.

\noindent\textbf{Analysis with respect to gradient.}
The gradient with respect to $e_k$ is
\begin{equation}
\frac{\partial \mathcal{L}_{\mathrm{SB}}}{\partial e_k}
=
\beta(e_k+\epsilon)^{\beta-1}
\end{equation}
Since $\beta<1$, directions with smaller reconstruction errors receive relatively larger gradient coefficients, while directions with large errors are prevented from dominating the optimization. Here, $\epsilon$ controls the maximum gradient coefficient. Similarly, the gradient with respect to the student coefficient can be written as \begin{equation} \frac{\partial \mathcal{L}_{\mathrm{SB}}} {\partial \tilde{c}_{b,k}^{S}} = (e_k+\epsilon)^{\beta-1}  \frac{\partial \mathcal{L}_{\mathrm{FM}}} {\partial \tilde{c}_{b,k}^{S}}. \end{equation} This shows that \ours can be interpreted as an adaptive spectral reweighting strategy, where each principal direction receives a weight determined by its reconstruction error. Thus, compared with vanilla feature matching, the proposed objective reduces the variance-dependent learning bias to approximately $\lambda_k^{\beta-1}$. Since $0<\beta<1$, the gap between high- and low-variance directions is substantially reduced. In summary, conventional feature matching tends to reproduce dominant spectral directions first, while under-optimizing non-dominant directions that may be important for downstream transfer. \ours counteracts this bias by balancing learning progress across teacher spectral directions, enabling the student to inherit richer representations. 

\noindent\textbf{Alternative formulation with static spectral balancing.}
Our analysis reveals that vanilla feature matching induces an optimization bias across spectral directions: principal directions with larger teacher variance tend to produce larger reconstruction errors and dominate the optimization. This observation suggests that the identified bias can be mitigated by explicitly accounting for the teacher spectrum.
One possible formulation is to directly assign a static weight to each principal direction according to its variance and leading to the following objective:
\begin{equation}
    \mathcal{L}_{\mathrm{static}}
    =
    \sum_{k=1}^{d}
    (\lambda_k+\epsilon)^{\beta-1} e_k.
    \label{eq:static_loss}
\end{equation}
Since $0<\beta<1$, this formulation suppresses the excessive contribution of high-variance directions while relatively emphasizing low-variance directions. Importantly, this objective follows directly from our analysis of the variance-dependent optimization bias and therefore represents another way of exploiting the proposed spectral-balancing principle. We provide the empirical analysis of this objective in Sec.~\ref{sec:additional_results} and confirm the superiority over the vanilla feature matching.

\section{Experiments}
\vspace{-2mm}
\subsection{Evaluation on Diverse Downstream Tasks}\label{subsec:exp_downstream}

We first evaluate the proposed method across diverse image classification datasets to assess its general applicability. For each dataset, we train a student to match the teacher’s representations and evaluate the resulting model on that dataset.

\noindent\textbf{Datasets.} 
We consider 14 benchmark datasets spanning seven categories: general image classification (CIFAR10 and CIFAR100~\citep{cifar}), fine-grained recognition (CUB~\citep{cub}), medical image classification (\citep{medmnistv2},\citep{bandi2018camelyon17}), remote sensing (\citep{resisc45}, \citep{helber2019eurosat}), anomaly detection (\citep{zou2022visa}, \citep{bergmann2019mvtec}), and domain generalization (\citep{iwildcam}, \citep{fmow}). 

\noindent\textbf{Models.} We use PE-Core-G14~\citep{bolya2026perception} and DINO-V2-ViT-G14 \citep{oquab2023dinov2} as the teacher model and evaluate two student architectures with different capacities: MobileNetV3-Small-1.0~\citep{howard2019searching} pre-trained on ImageNet~\citep{deng2009imagenet} and PE-Core-T14~\citep{bolya2026perception} pre-trained on image-text data. 

\newcommand{\datatag}[1]{\rotatebox[origin=l]{90}{\scriptsize{#1}}}
\definecolor{lightgreen}{HTML}{D8ECD1}
\definecolor{lightgray}{gray}{0.9} 
\definecolor{bettergreen}{RGB}{35,140,70}
\definecolor{worsered}{RGB}{190,60,60}

\newcommand{\better}[1]{{{\textbf{#1}}}}
\newcommand{\worse}[1]{\textcolor{worsered}{{\textbf{#1}}}}

\begin{table*}[t]
\centering
\small
\caption{
\small
Performance comparison across 14 datasets, including general recognition, OCR, fine-grained recognition, medical image classification, and anomaly detection benchmarks. The cell where the distillation model underperforms the student linear probe is highlighted with \worse{red} while the best model is highlighted with \better{bold}.
}
\vspace{-3mm}
\setlength{\tabcolsep}{3pt}
\resizebox{1.0\linewidth}{!}{
\begin{tabular}{l c | c c   | c c | c c |c c| c c | c c| c c}
\toprule
&
&
\multicolumn{2}{c|}{\scriptsize General}
&
\multicolumn{2}{c|}{\scriptsize OCR}
&
\multicolumn{2}{c|}{\scriptsize Fine}
&
\multicolumn{2}{c|}{\scriptsize Remote}
&
\multicolumn{2}{c|}{\scriptsize Medical}
&
\multicolumn{2}{c|}{\scriptsize Anomaly}
&
\multicolumn{2}{c}{\scriptsize Generalization}
\\
&
\datatag{Average}
&
 \datatag{CIFAR100}
& \datatag{CIFAR10}
& \datatag{SVHN}
& \datatag{GTSRB}
& \datatag{NABirds}
& \datatag{CUB}
& \datatag{Resisc}
& \datatag{Eurosat}
& \datatag{OrganCMNIST}
& \datatag{Camelyon17}
& \datatag{VisA}
& \datatag{MVTec}
& \datatag{iWildCam}
& \datatag{FmoW}
\\
& \scriptsize \# classes
& \scriptsize 100 & \scriptsize 10
& \scriptsize 10  & \scriptsize 43
& \scriptsize 555 & \scriptsize 200
& \scriptsize 45  & \scriptsize 10
& \scriptsize 11  & \scriptsize 2
& \scriptsize 2   & \scriptsize 2
& \scriptsize 186 & \scriptsize 62
\\ \hline
\rowcolor{blue!10} PE-Core-Large $\rightarrow$ PE-Core-Tiny &  & & & &  & & & & & & & & & &  \\
Teacher
&86.5 &93.1&99.4&	73.6&93.7&87.3&	90.6&96.1&97.4&	82.9&90.1&81.9&96.3		&79.2&50.1
\\
Student 
& 
76.0 
& 76.2&92.1&57.7&81.5&59.7&71.1&88.9&94.6&82.9&89.0&75.8&86.3&71.0&36.5\\
Logits-KD~\citep{hinton2015distilling} &80.9 &79.5 & 95.1&		82.2&96.8&69.9&71.6&93.2&	98.4&92.4&\worse{88.8}&\worse{69.2}& \worse{71.9}&74.1& 45.3\\\arrayrulecolor{gray}
\midrule
DINO~\citep{caron2021emerging}
& 
79.3 
& 
\worse{74.2}&94.3	&85.5&89.9&69.2&72.7&91.5&97.8&85.9&91.1&\worse{62.9}&89.4&\worse{66.2}&40.4\\ 

RKD~\citep{park2019relational}&\worse{70.0}&\worse{60.7} & 93.7  &  67.3&90.8&62.3	&76.1&\worse{19.2} & \worse{92.7}&84.1&91.5&\worse{66.1}&92.4&74.3&\worse{9.3} \\ 
FM
& 
83.0& 

80.9&97.1&84.1&93.3&70.3&77.5&93.6&98.3&83.8&\better{94.0}&\worse{74.2}&96.5&71.7&43.6
\\
\ours
& \better{{84.6}} 
&\better{81.9}&\better{97.3}&\better{85.8}&\better{94.8}&\better{73.7}&\better{79.6}&\better{94.8}&\better{98.4}&\better{89.2}&93.4&\better{76.6}&\better{97.9}	&\better{74.5}&\better{47.4}\\\hline
\rowcolor{blue!10} PE-Core-Large $\rightarrow$ MobileNet-V3 &   & & & & & & & & & & & & & & \\
Teacher
&86.5 &93.1&99.4&	73.6&93.7&87.3&	90.6&96.1&97.4&	82.9&90.1&81.9&		96.3&79.2&50.1
\\
Student &76.0&
64.6&86.5&	65.0&77.6&55.4&	67.6&86.3&94.9&	85.2&85.7&72.8&84.6&		58.6&26.1\\
Logits-KD~\citep{hinton2015distilling} & 80.3&76.3&94.2	&92.9&96.6&	60.8&67.9&93.0&98.1&92.6&88.8&75.1&89.2&\worse{57.8}&40.2\\\arrayrulecolor{gray}
\midrule
DINO~\citep{caron2021emerging} &79.3&74.8
&94.3&87.4&	93.3&\worse{54.7}&\worse{66.6}&	91.5&97.8&88.4&	93.0&76.5&96.0&		59.7&35.7\\ 
RKD~\citep{park2019relational} &\worse{70.8}& 70.0&\worse{92.0}& 81.3 & 89.8 & 	\worse{54.7} & 70.0	 &89.7&97.4 &88.4&91.4&76.8&93.7& \worse{56.3} & 32.7\\ 
FM 
& 
80.9& 
76.0&94.8&87.6&93.9&	62.6&71.3&92.6&\better{98.2}&89.2&	93.0&80.7&95.5&\worse{58.5}&39.0\\

\ours
 
& \better{82.4}
&\better{77.6}&\better{95.2}&\better{89.5}&\better{94.6}&	\better{64.2}&\better{72.5}&\better{93.6}&\better{98.2}&\better{90.7}&\better{93.9}&\better{81.9}&\better{96.8}&\better{59.9}&\better{42.1}\\\hline

\rowcolor{blue!10} DINO-V2 $\rightarrow$ MobileNet-V3 &   & & & & & & & & & & & & & & \\
Teacher &84.0 & 92.7 & 99.3	 & 62.7 & 77.6 & 87.8 & 90.3 & 93.8 & 96.7&87.6		&87.1&82.1&95.2&81.1&41.4\\

Student &76.0& 64.6 & 86.5	& 	65.0	& 77.6	& 55.4	& 67.6& 	86.3& 	94.9	&85.2 &	85.7	& 72.8	& 84.6		&58.6& 26.1\\
Logits-KD~\citep{hinton2015distilling} &80.1& 76.7 & 94.3 & 88.7&95.6&61.0&67.9&92.9&98.3&93.5&92.0&77.0&86.4&\worse{57.7}&39.0\\\arrayrulecolor{gray}
\midrule
DINO~\citep{caron2021emerging} &76.9& 75.5&94.4&78.2&84.5&\worse{52.6}&\worse{66.3}&91.7&97.5&88.8 & 86.1 & 74.4 & 91.3&\better{61.2}&35.3 \\ 
RKD~\citep{park2019relational} &\worse{75.8}& 69.4 & 92.0 & 77.0 & 83.0 &	\worse{54.7}&\worse{64.3}&89.9&96.9&87.2&89.5&74.9&91.7&\worse{58.3}&32.1\\
FM &78.7 & 77.2 & 95.2& 80.5 & 87.7 & \worse{53.6} & \worse{66.4} & 92.6 & 	97.9 &89.6& 93.2 & 76.6 & 93.5&60.2	 & 37.5\\
\ours & \better{79.8} &  \better{78.1} & \better{95.5} & \better{82.3} & \better{88.1} & \better{56.8} & \better{69.0} & \better{92.8} & \better{98.0}&	\better{91.0}	&\better{93.4} & \better{77.8} & \better{94.3}	&61.1& \better{39.5}\\ 

\bottomrule
\end{tabular}
}
\label{tab:main_results}
\end{table*}

\noindent\textbf{Baselines.} 
Relational knowledge distillation (RKD)~\citep{park2019relational}, feature matching with L2-distance (FM), and DINO~\citep{caron2021emerging} are used as the baseline for the model distillation with unlabeled data. The implementation details are available in Sec.~\ref{sec:exp_details}. As a reference to the distillation baseline using labeled data during training, we additionally report Logits-KD~\citep{hinton2015distilling}, where a classification head is trained on the teacher using labeled data, and the resulting logits are used as soft targets for students. 

\noindent\textbf{Training.} 
All models are trained with AdamW~\citep{loshchilov2019decoupled}. Since FM, DINO and \ours need to train the linear head, we freeze the base model for 500 iterations and tune all parameters. $\beta^{-1}$ is set as $3$ in all datasets. Other details are provided in Sec.~\ref{sec:exp_details}. 

\noindent\textbf{Evaluation.} After unlabeled feature matching, the learned representations are evaluated by linear-probing and report classification accuracy. For MVTec, we compute the distance to the nearest neighbor to compute AUROC. We report the performance averaged over three runs. 

\noindent\textbf{Overview of the results.} 
According to Table~\ref{tab:main_results}, \ours outperforms all baselines in almost every setting (40/42 settings), demonstrating the effectiveness of distilling rich feature representations. Notably, the performance gains are consistent across all combinations of teacher pre-training strategies and student architectures, indicating that our method generalizes well across diverse distillation settings. Several observations can be drawn from these results. 

\noindent\textbf{\ours consistently improves the pretrained student.} First, while several distillation baselines even degrade performance compared to the pretrained student, as highlighted by red numbers, \ours consistently improves performance in all evaluated settings. Transferring rich feature representations is a highly effective strategy for improving downstream task performance. 

\noindent\textbf{\ours provides larger gains on fine-grained recognition.}
Second, the improvement over the feature matching (FM) baseline is relatively modest on general recognition tasks such as CIFAR-10 and CIFAR-100, whereas substantially larger gains are observed on fine-grained datasets such as CUB and NABirds. This trend is consistent with the analysis in Figure~\ref{fig:motivation}, which shows that general recognition primarily relies on high-variance feature directions, while fine-grained recognition benefits more from preserving lower-variance components. 


\noindent\textbf{\ours is effective for domain generalization.}
\ours yields large improvements on domain generalization benchmarks. We conjecture that preserving richer representations leads to more transferable and robust representations, thereby improving generalization to unseen domains. 

\noindent\textbf{The student can even surpass the teacher on out-of-domain tasks.}
Finally, the student frequently surpasses the teacher on several domains, particularly OCR, medical imaging, and remote sensing, even with FM. These domains are likely underrepresented in the teacher's pretraining data, suggesting that our distillation strategy enables the student to leverage the teacher's rich representations while adapting them more effectively to out-of-domain tasks.

\begin{figure}[t]
    \centering

    \begin{subfigure}{0.4\linewidth}
        \centering
        \includegraphics[width=\linewidth]{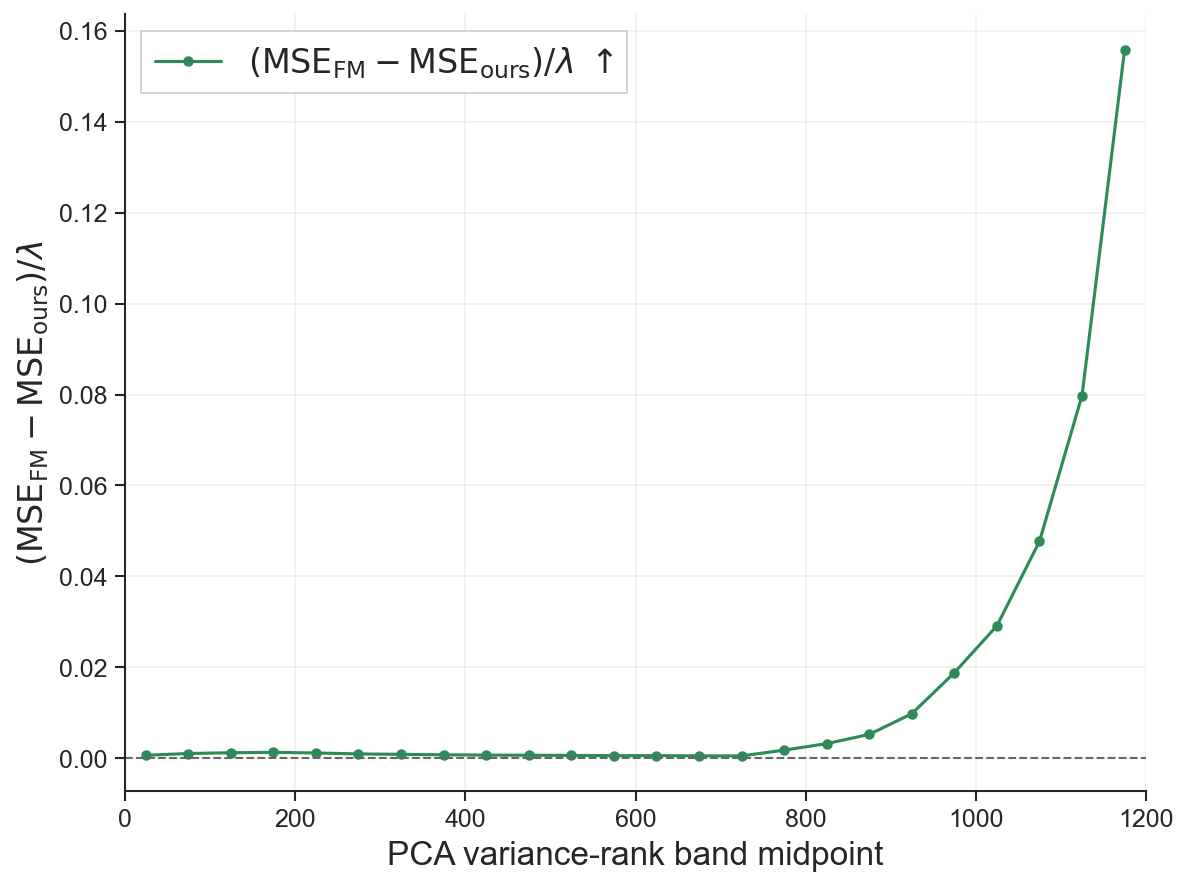}
        \caption{CIFAR100}
        \label{fig:cifar100}
    \end{subfigure}
   \hspace{0.06\linewidth}
    \begin{subfigure}{0.4\linewidth}
        \centering
        \includegraphics[width=\linewidth]{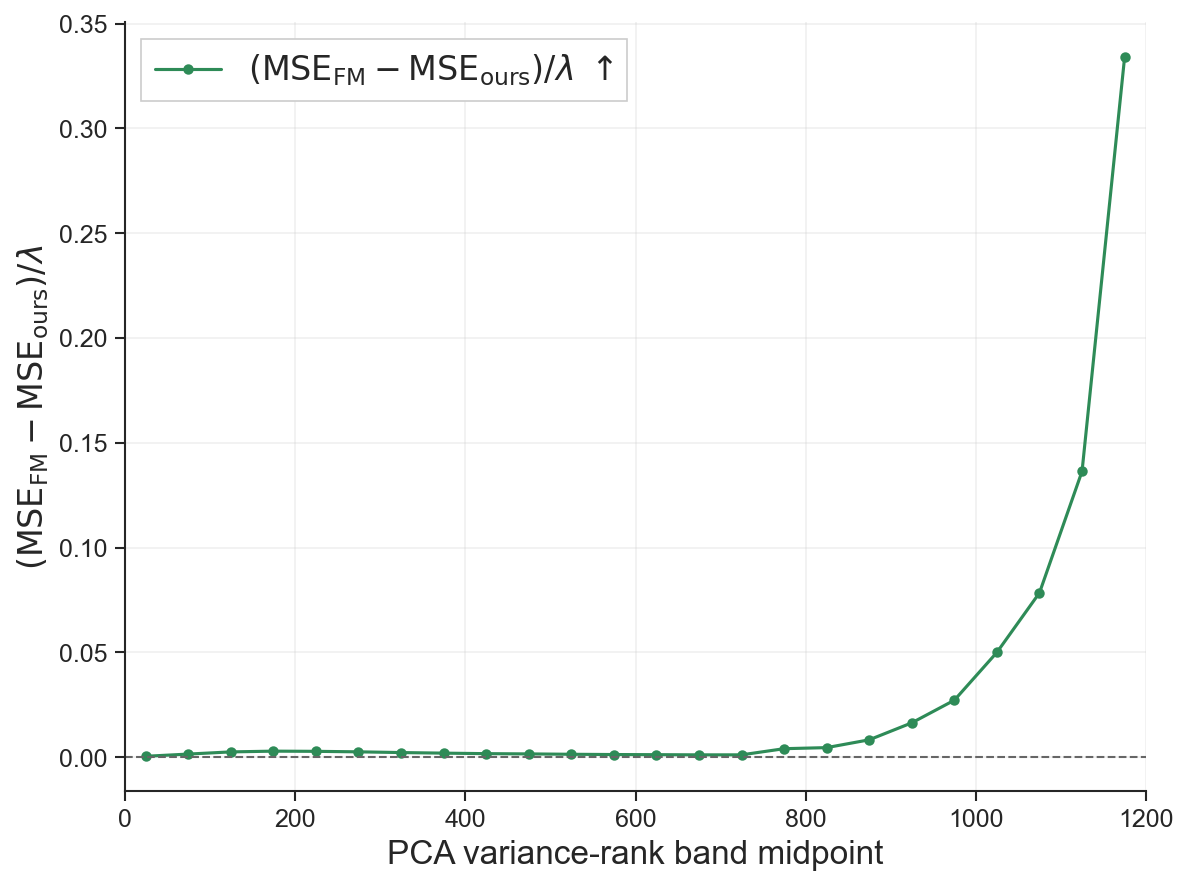}
        \caption{NABirds}
        \label{fig:nabirds_mse}
    \end{subfigure}
\vspace{-3mm}
    \caption{\small
    \textbf{Difference in reconstruction errors across the principal components.} We compute the mean reconstruction error between the teacher representation and the student, and report ${{e}^{\mathrm{FM}}_{k}} - {{e}^{\mathrm{ours}}_{k}}$ normalized by the teacher variance $\lambda_{k}$. A higher value shows that \ours reconstructs the teacher representation more accurately than the feature matching baseline.}
    \label{fig:mse_diff}
\end{figure}
\begin{table*}[t]
\vspace{-3mm}
\centering
\caption{\small
Experiments on distillation with ImageNet.
All models are evaluated with classification accuracy (\%) after linear-probing on the corresponding labeled subset.}
\vspace{-3mm}
\resizebox{0.8\textwidth}{!}{
\begin{tabular}{lccc|c|ccc}
\toprule
Teacher&Student & Params &Initialization  & Method
& 5-shot & 10-shot & Full-shot \\
\midrule
\multirow{3}{*}{PE-Core-G14}& \multirow{3}{*}{ViT-Tiny}&\multirow{3}{*}{5.5M}& \multirow{3}{*}{Scratch}    & DINO & 31.3&37.2&54.6\\
&&&&FM      & 48.4 & 51.6 & 57.6\\
& 
&  & & 
\ours            & \textbf{50.2} & \textbf{52.7} & \textbf{58.4}  \\
\midrule
\multirow{3}{*}{ConvNEXT-XLarge} & \multirow{3}{*}{ConvNEXT-Tiny}  & \multirow{3}{*}{27.8M}& \multirow{3}{*}{Scratch}&DINO&36.4&42.9&54.7\\
&&&&FM & 57.6 & 60.6&67.1\\
& &    &&
\ours&  \textbf{61.2} & \textbf{63.9} & \textbf{70.0}\\
\midrule
\multirow{4}{*}{PE-Core-G14} & \multirow{4}{*}{PE-Core-T14}  &\multirow{4}{*}{6.1M}& \multirow{4}{*}{Pretrained}&Base & 45.3 & 52.1 & 68.1\\
&&    &  & DINO & 60.0&62.6& 68.3\\
&&    &  & FM      & 59.3 & 63.5 & 72.1 \\
& &    &&  \ours            & \textbf{60.3} & \textbf{64.4} & \textbf{73.1}  \\
\bottomrule
\end{tabular}}

\label{tab:imagenet_shots}
\end{table*}

\subsection{Analysis.} 
\vspace{-2mm}

\noindent\textbf{Reconstruction Error in each basis.} 
Figure~\ref{fig:mse_diff} compares the reconstruction errors of Feature Matching (FM) and \ours along each principal direction. Let ${{e}^{\mathrm{FM}}_{k}}$ and ${{e}^{\mathrm{ours}}_{k}}$ denote the mean squared reconstruction errors along the $k$-th principal direction for FM and \ours, respectively. We visualize their difference, ${{e}^{\mathrm{FM}}_{k}} - {{e}^{\mathrm{ours}}_{k}}$, normalized by the teacher variance $\lambda_k$ along the corresponding direction.
Positive values indicate that \ours reconstructs the teacher representations more accurately than FM. \ours consistently achieves lower reconstruction error across the entire spectrum, demonstrating that the improvement is not limited to a specific subset of principal components. Notably, the gain becomes substantially larger for low-variance components, indicating that our objective effectively addresses the optimization imbalance highlighted in Fig.~\ref{fig:motivation}, where conventional feature matching tends to underfit low-variance directions. This observation further suggests that improving the reconstruction of these previously under-learned directions contributes directly to the superior downstream performance. Interestingly, although our objective balances optimization across spectral directions, it also yields a small but consistent improvement in high-variance components. This suggests that balancing the optimization landscape not only prevents the neglect of low-variance directions but also facilitates overall optimization, leading to better reconstruction even in the dominant spectral components.

\noindent\textbf{Experiments on ImageNet.} Table~\ref{tab:imagenet_shots} presents analysis on ImageNet. We train a student model from scratch or fine-tune a pre-trained student model. In both cases, the models are trained without labels. Our objective shows consistent improvement over vanilla feature matching. These results indicate that the proposed method remains effective even on a large-scale dataset such as ImageNet. Furthermore, the improvements are consistently observed under both 5-shot and 10-shot linear evaluation protocols.

\begin{figure}[t]
    \centering
    \begin{subfigure}{0.3\linewidth}
        \centering
        \includegraphics[width=\linewidth]{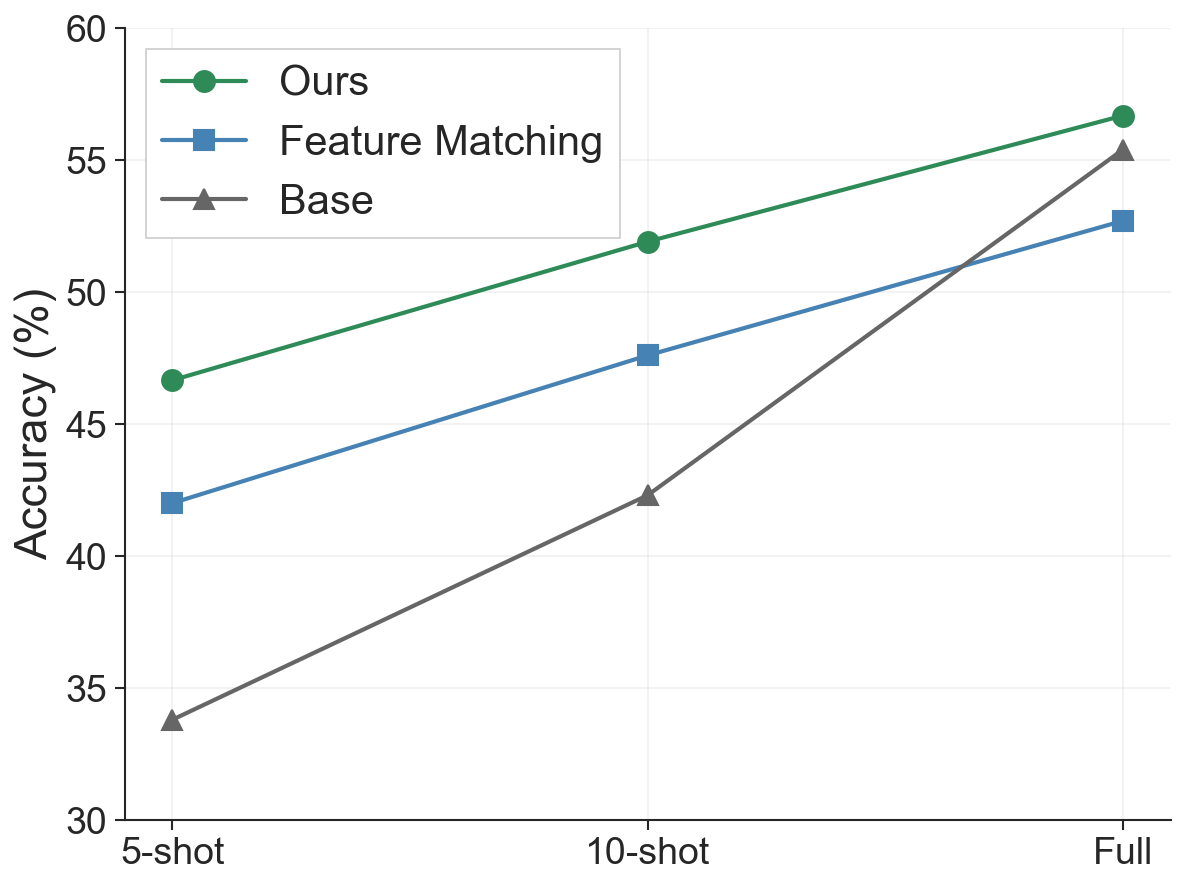}
        \caption{NABirds}
        \label{fig:nabirds}
    \end{subfigure}
    \hfill
    \begin{subfigure}{0.3\linewidth}
        \centering
        \includegraphics[width=\linewidth]{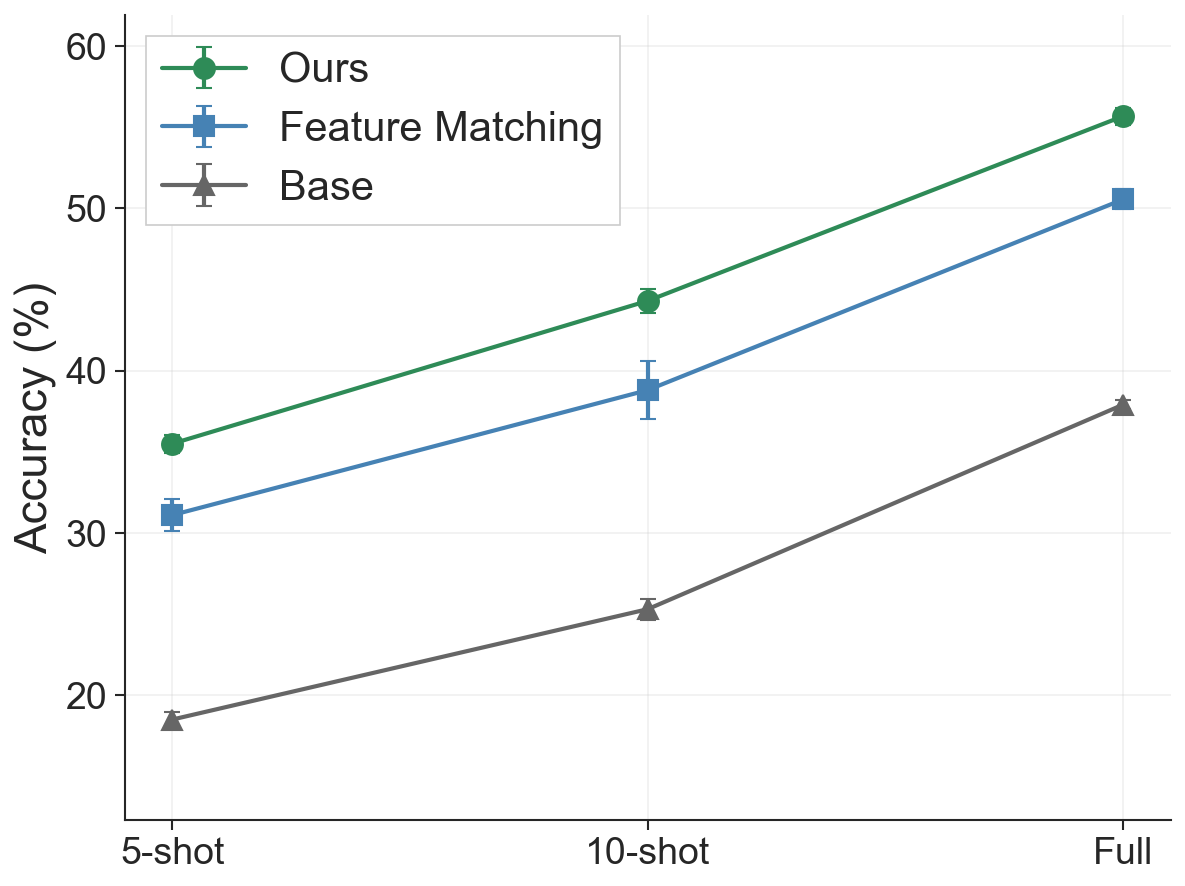}
        \caption{FGVC}
        \label{fig:fgvc}
    \end{subfigure}
    \hfill
    \begin{subfigure}{0.3\linewidth}
        \centering
        \includegraphics[width=\linewidth]{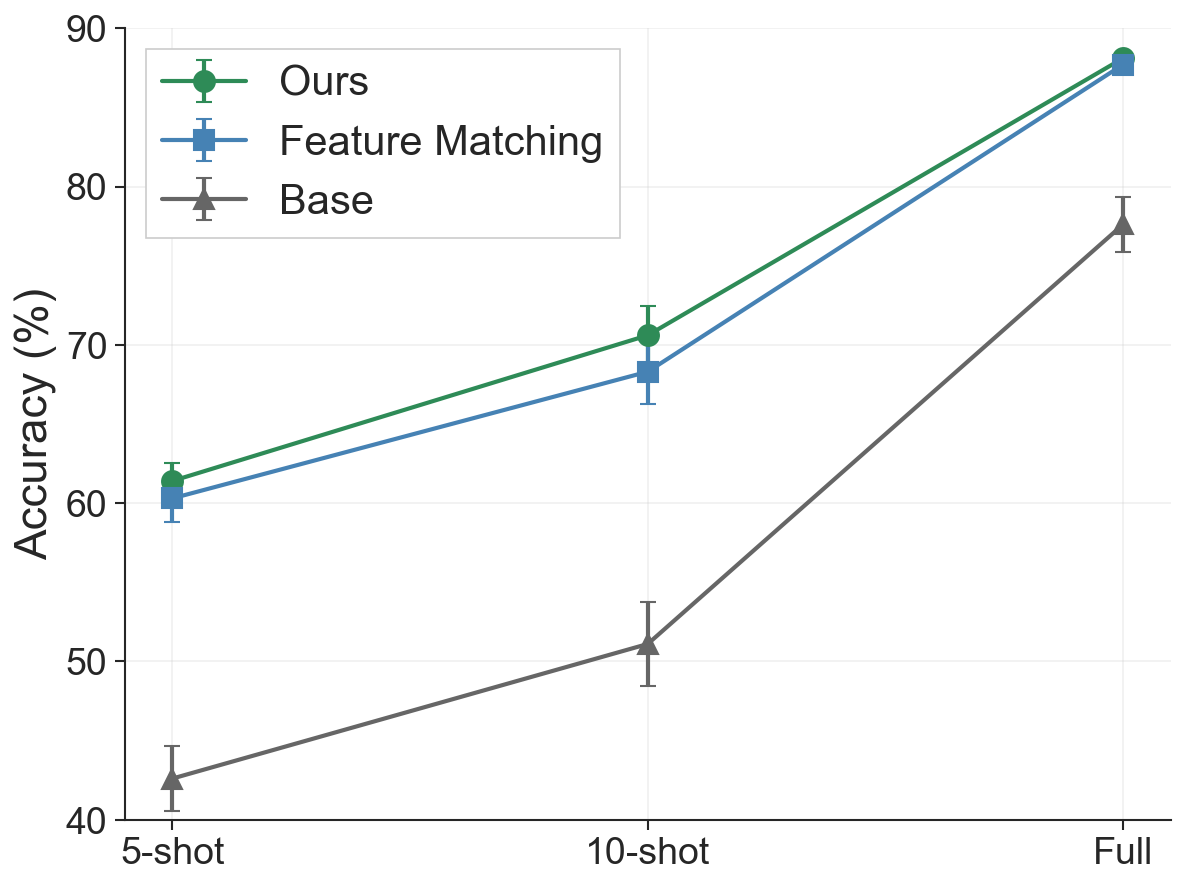}
        \caption{GTSRB}
        \label{fig:voc}
    \end{subfigure}
    \vspace{-3mm}
    \caption{\small Linear probing with different shots. Our approach consistently improves performance over the base student model while vanilla feature matching sometimes fails in improving as in NABirds full-shot case (a). }
    \label{fig:fewshot}
\end{figure}

\begin{table*}[t]
\scriptsize
\setlength{\tabcolsep}{2.0pt}
\centering
\caption{
\small
\textbf{\figleft:} Sensitivity analysis for $\beta^{-1}$. $\beta=1$ corresponds to vanilla feature matching, while $\beta^{-1}=2$ and $3$ consistently improve performance.
\textbf{\figright:} Comparison among different objectives, including L1, cosine distance, and L2 + whitening. \ours is the best or the second best approach in these results.  
}
\resizebox{0.9\textwidth}{!}{
\begin{tabular}{lcccc}
\toprule
$\beta^{-1}$ & ImageNet & NAB & CIF100 & SVHN\\
\midrule
1 (FM) & 72.1 &60.7 & 76.2& 87.2 \\
2 & 72.7 & 62.3 & 77.5 & 89.4 \\
3 & \textbf{73.1} & 64.2 & 77.6 & 89.5 \\
4 & 72.9 & 64.9 & 77.8 & 89.9 \\
5 &72.9&64.9&78.2&89.4\\
\bottomrule
\end{tabular}
\hspace{5mm}
\begin{tabular}{lccccccc}
\toprule
Objective & NAB & CIF100 & Cam17 & GTSRB & SVHN & FmoW\\
\midrule
L1 & 63.6 & 76.4&\textbf{94.3}&93.9&88.9&38.8\\
Cosine & 61.3 & 75.8&93.2&92.4&87.1&37.3\\
L2 & 62.6&76.0&93.0&93.9&87.6&39.0\\
L2 + Whitening & 63.2 & \textbf{77.8}&93.1& 
91.6&86.7&39.4\\
\rowcolor{ourscolor}\ours & \textbf{64.8} & 77.6 & 93.9&\textbf{94.6}&\textbf{89.5}&\textbf{42.1}\\
\bottomrule
\end{tabular}
}

\label{tab:ablation}

\end{table*}

\noindent\textbf{Few-shot probing.} Figure~\ref{fig:fewshot} presents few-shot linear probing results on fine-grained datasets (NABirds and FGVC Aircraft~\citep{maji2013fine}) and GTSRB. Compared with vanilla feature matching, our method consistently improves performance across all datasets and shot settings. The gains are particularly pronounced on the fine-grained datasets, suggesting that balancing feature learning across spectral directions is especially beneficial for tasks requiring subtle visual discrimination. Moreover, while the distilled representations consistently outperform the original student representations, the improvements are larger in the few-shot regime than in the full-shot setting. As more labeled data becomes available, the performance gap gradually narrows.

\noindent\textbf{Analysis of the distance function.} 
Previous feature matching methods have employed L1, L2, or cosine distance. We therefore compare these objectives in the right part of Table~\ref{tab:ablation}. \ours achieves either the best or second-best performance on all datasets, yielding the strongest performance on average. In contrast, none of the conventional distance functions consistently performs well across datasets: each can be effective on certain tasks but performs relatively poorly on others. These results demonstrate the robustness of \ours across diverse downstream tasks.

\noindent\textbf{Does whitening target features improve performance?}
Given the discussion in Sec.~\ref{sec:motivation_insight}, a natural alternative for balancing the training signal is to whiten the target features, \ie normalizing each spectral direction by its variance, as studied by \cite{ranzinger2024phi}. We therefore train the student to predict whitened teacher features after PCA projection using an MSE loss. The right of Table~\ref{tab:ablation} shows that whitening does not necessarily improve the performance over the L2 distance. We hypothesize that this is because whitening completely equalizes the training signal, disregarding the relative importance of high-variance directions that may capture prominent variations in the data. In contrast, \ours reduces the imbalance across spectral directions while preserving their relative importance, which we find more effective for downstream performance.

\noindent\textbf{Hyper-parameter sensitivity.} The left part of Table~\ref{tab:ablation} studies the effect of $\beta$. Increasing $\beta^{-1}$ from 1 progressively balances the contributions of different spectral directions. All tested values improve performance over the unweighted baseline, while the differences among them remain relatively small. Thus, the proposed method is not highly sensitive to $\beta$ within the evaluated range.

\begin{table}[t]
    \centering
\setlength{\tabcolsep}{4.5pt}
    \caption{
    \small
    Study on imbalanced data distribution using PE-Core-G14 and PE-Core-Tiny as teacher and student, respectively.
    \textbf{\figleft: }Anomaly detection performance (AUROC) on ViSA with different numbers of anomaly samples used for distillation.
    \textbf{\figright: }Results on iNaturalist2018 across head, medium, and tail classes. 
    }
     \vspace{-3mm}
    \resizebox{0.75\linewidth}{!}{
    \begin{minipage}[t]{0.4\linewidth}
        \centering
        \vspace{0pt}
        \small
        \renewcommand{\arraystretch}{1.1}
        \begin{tabular}{lcccc}
        \toprule
        \multirow{2}{*}{Method}
        & \multicolumn{4}{c}{ViSA \# Anomaly Samples} \\
        \cmidrule(lr){2-5}
        & 100 & 200 & 500 & 1000 \\
        \midrule
         Teacher  & \multicolumn{4}{c}{0.898} \\
        Student & \multicolumn{4}{c}{0.825} \\\hline
       
        FM      & 0.835 & 0.851 & 0.863 & 0.880 \\
       \rowcolor{ourscolor} \ours    & \textbf{0.851} & \textbf{0.867} & \textbf{0.873} & \textbf{0.896} \\
        \bottomrule
        \end{tabular}
    \end{minipage}
    \hspace{0.08\linewidth}
    \begin{minipage}[t]{0.4\linewidth}
        \centering
        \vspace{0pt}
        \small
        \renewcommand{\arraystretch}{1.1}
        \begin{tabular}{lcccc}
        \toprule
        \multirow{2}{*}{Method}
        & \multicolumn{4}{c}{iNaturalist2018} \\
        \cmidrule(lr){2-5}
        & Head & Medium & Tail & All \\
        \midrule
        Teacher  & 86.7 & 78.0 & 73.7 & 77.2 \\
        Student & 54.2 & 47.2 & 44.5 & 46.8 \\\hline
        FM           & 60.3 & 51.8 & 51.0 & 52.3 \\
       \rowcolor{ourscolor} \ours         & \textbf{61.3} & \textbf{52.7} & \textbf{52.0} & \textbf{53.3} \\
        \bottomrule
        \end{tabular}
    \end{minipage}
    }
    \label{tab:anomaly_inat}
\end{table}

\begin{table*}[t]
\centering
\small

\caption{\small
Protein downstream performance on six PFMBench tasks.
Results are scaled by 100 and reported as mean $\pm$ standard deviation over three runs. Higher is better for all metrics.
}
\vspace{-3mm}
\resizebox{1.0\linewidth}{!}{
\setlength{\tabcolsep}{4pt}
\begin{tabular}{l|cc|cc|cc|cc|cc|cc}
\toprule
& \multicolumn{2}{c|}{Fold: 1195 classes}
& \multicolumn{2}{c|}{GO-MF: 489 classes}
& \multicolumn{2}{c|}{GO-CC: 320 classes}
& \multicolumn{2}{c|}{EC: 585 classes}
& \multicolumn{2}{c|}{Cat. Eff. : regression}
& \multicolumn{2}{c}{DeepLoc2: 10 classes} \\
Method
& Val & Test
& Val & Test
& Val & Test
& Val & Test
& Val & Test
& Val & Test \\
\midrule

Teacher (ESM-2 650M) & $75.0_{\pm 0.1}$ & $73.1_{\pm 0.3}$ & $55.6_{\pm 0.2}$ & $55.5_{\pm 0.4}$ & $45.7_{\pm 0.4}$ & $47.6_{\pm 0.2}$ & $58.8_{\pm 0.2}$ & $59.8_{\pm 0.2}$ & $7.5_{\pm 0.4}$ & $10.6_{\pm 0.5}$ & $71.7_{\pm 0.5}$ & $70.9_{\pm 0.1}$ \\ 

Student (ESM-2 6M) & $49.4_{\pm 0.3}$ & $51.2_{\pm 0.2}$ & $34.5_{\pm 0.4}$ & $36.6_{\pm 0.3}$ & $30.4_{\pm 0.5}$ & $29.7_{\pm 0.2}$ & $38.7_{\pm 0.1}$ & $40.1_{\pm 0.3}$ & $14.3_{\pm 0.3}$ & $17.0_{\pm 0.5}$ & $64.3_{\pm 0.1}$ & $62.9_{\pm 0.1}$ \\

\midrule

L1
& $52.6_{\pm 0.3}$ & $53.1_{\pm 0.3}$
& $36.5_{\pm 0.1}$ &  $37.5_{\pm 0.2}$
& $35.2_{\pm 0.1}$ & $34.8_{\pm 0.1}$
& $38.3_{\pm 0.2}$ & $40.9_{\pm 0.4}$
& $20.7_{\pm 1.3}$ & $22.0_{\pm 0.8}$
& $\mathbf{70.1}_{\pm 0.1}$ & $68.4_{\pm 0.1}$ \\

L2
& $51.9_{\pm 0.2}$ & $52.2_{\pm 0.3}$
& $36.8_{\pm 0.2}$ &  $38.0_{\pm 0.2}$
& $35.2_{\pm 0.2}$ & $35.3_{\pm 0.3}$
& $39.9_{\pm 0.4}$ & $41.0_{\pm 0.4}$
& $18.8_{\pm 1.3}$ & $27.0_{\pm 0.8}$
& $69.6_{\pm 0.1}$ & $68.5_{\pm 0.1}$ \\

\rowcolor{ourscolor}
\ours
& $\mathbf{54.4}_{\pm 0.6}$ & $\mathbf{55.6}_{\pm 0.5}$
& $\mathbf{38.6}_{\pm 0.2}$ & $\mathbf{39.6}_{\pm 0.1}$ 
& $\mathbf{35.8}_{\pm 0.2}$ & $\mathbf{35.8}_{\pm 0.2}$
& $\mathbf{41.7}_{\pm 0.3}$ & $\mathbf{42.0}_{\pm 0.6}$
& $\mathbf{21.3}_{\pm 0.6}$ & $\mathbf{30.7}_{\pm 1.0}$
& $69.9_{\pm 0.3}$ & $\mathbf{69.2}_{\pm 0.1}$ \\
\bottomrule
\end{tabular}
}

\label{tab:protein_results}
\end{table*}

\noindent\textbf{Robustness to imbalanced class distribution.} 
Table~\ref{tab:anomaly_inat} analyzes the effectiveness of \ours under imbalanced unlabeled data distributions. On the left, we progressively reduce the number of unlabeled anomaly samples used during distillation. As the number of anomaly samples decreases, the overall performance gradually drops, likely because anomaly-specific feature patterns are observed less frequently during training and therefore receive fewer optimization updates. Nevertheless, \ours consistently outperforms vanilla feature matching across all settings, demonstrating its robustness even when anomaly samples are scarce. On the right of Table~\ref{tab:anomaly_inat}, we assess on iNaturalist2018~\citep{van2018inaturalist}, a long-tailed recognition dataset with 8,142 categories. \ours consistently improves over the baseline on all metrics. These indicate that the proposed objective remains effective under highly imbalanced class distributions and learns more discriminative representations for both well-represented and under-represented classes.

\noindent\textbf{Experiments on protein embedding model.} 
\ours also generalizes to protein foundation models. 
We distill ESM-2 650M into ESM-2 6M~\citep{lin2023evolutionary} and evaluate the learned representations via linear probing on six PFMBench tasks~\citep{gao2025pfmbench}. Fold denotes protein fold classification on the Remote Homology benchmark~\citep{lo2000scop}. GO-MF and GO-CC are multi-label Gene Ontology prediction tasks for molecular function and cellular component, respectively~\citep{ashburner2000gene}. EC denotes multi-label enzyme commission number prediction~\citep{bairoch2000enzyme}. Cat. Eff. measures enzyme catalytic efficiency prediction using Spearman correlation~\citep{li2022deep}, while DeepLoc2 evaluates multi-label protein subcellular localization prediction~\citep{thumuluri2022deeploc}. As in the image experiments, distillation is performed separately on the training set of each task, and evaluation follows the official protocol and metrics. The consistent improvements across these diverse tasks suggest that useful information is distributed beyond high-variance components in both image and protein representations. Thus, spectral balancing provides a robust feature-distillation objective that generalizes across modalities.


\vspace{-2mm}
\section{Conclusion}
\vspace{-3mm}
We presented Spectrum-Balanced Feature Matching, \ours, a simple and effective objective for feature distillation that addresses the imbalance of optimization across spectral directions. By dynamically reweighting the reconstruction loss according to the reconstruction error of each principal component, the proposed objective encourages the student to learn richer feature representations while introducing negligible computational overhead. Extensive experiments on diverse downstream tasks demonstrate that \ours consistently improves over conventional feature matching across a wide range of teacher--student combinations and datasets. We hope that this work provides a simple yet effective direction for improving feature distillation.

\subsection*{Acknowledgement}
This research was supported by JST PRESTO, Japan, Grant Number JPMJPR2523. This work was partly achieved through the use of SQUID at D3 Center, The University of Osaka. This research was conducted using the Supermicro ARS-111GL-DNHR-LCC and FUJITSU Server PRIMERGY CX2550 M7 (Miyabi) at Joint Center for Advanced High Performance Computing (JCAHPC). This work was supported in part by the Physical AI Development Support Program by AWS Japan through the provision of computational resources.

\subsection*{AI use statement}
In this work, we used generative AI tools to assist with writing code and with manuscript writing/editing. All AI-assisted outputs were reviewed by the authors. In particular, AI-assisted code was inspected and verified by the authors, and AI-assisted text was edited and checked for accuracy, consistency with the experiments, and originality. We take responsibility for the final content of this work, including all text, claims, code, and artifacts produced with the aid of generative AI.

\subsection*{Ethics statement}
This work studies knowledge distillation and representation transfer using publicly available datasets and pretrained models. Our experiments do not involve human subjects or the collection of personally identifiable or sensitive information. We follow the licenses and intended research use of the datasets and models used in our experiments. We do not identify any direct ethical concerns specific to the proposed methodology beyond those generally associated with the underlying pretrained models and datasets.

\subsection*{Reproducibility statement}
The Pytorch-style code of the proposed objective is shown in Algorithm~\ref{alg:specmatch}. The overview of the experimental setup is described in Sec.~\ref{subsec:exp_downstream}. 
More specific details of the experiments, e.g., models and hyper-parameters, are described in the Sec.~\ref{sec:exp_details}. We will also release the code used for our experiments upon acceptance.



\bibliography{iclr2027_conference}

\begin{thebibliography}{50}
\providecommand{\natexlab}[1]{#1}
\providecommand{\url}[1]{\texttt{#1}}
\expandafter\ifx\csname urlstyle\endcsname\relax
  \providecommand{\doi}[1]{doi: #1}\else
  \providecommand{\doi}{doi: \begingroup \urlstyle{rm}\Url}\fi

\bibitem[Ashburner et~al.(2000)Ashburner, Ball, Blake, Botstein, Butler, Cherry, Davis, Dolinski, Dwight, Eppig, et~al.]{ashburner2000gene}
Michael Ashburner, Catherine~A Ball, Judith~A Blake, David Botstein, Heather Butler, J~Michael Cherry, Allan~P Davis, Kara Dolinski, Selina~S Dwight, Janan~T Eppig, et~al.
\newblock Gene ontology: tool for the unification of biology.
\newblock \emph{Nature genetics}, 25\penalty0 (1):\penalty0 25--29, 2000.

\bibitem[Bairoch(2000)]{bairoch2000enzyme}
Amos Bairoch.
\newblock The enzyme database in 2000.
\newblock \emph{Nucleic acids research}, 28\penalty0 (1):\penalty0 304--305, 2000.

\bibitem[Bandi et~al.(2019)Bandi, Geessink, Manson, Reiter, Balkenhol, Hermsen, Bejnordi, Bandettini~di Poggio, Rutgers, Litjens, and et~al.]{bandi2018camelyon17}
Peter Bandi, Oscar Geessink, Quirine Manson, Mayer Reiter, Marc Balkenhol, Meyke Hermsen, Babak~Ehteshami Bejnordi, Brenda Bandettini~di Poggio, Iris Rutgers, Geert Litjens, and et~al.
\newblock From detection of individual metastases to classification of lymph node status at the patient level: The camelyon17 challenge.
\newblock \emph{IEEE Transactions on Medical Imaging}, 38\penalty0 (2):\penalty0 550--560, 2019.
\newblock \doi{10.1109/TMI.2018.2867350}.

\bibitem[Beery et~al.(2018)Beery, Van~Horn, and Perona]{iwildcam}
Sara Beery, Grant Van~Horn, and Pietro Perona.
\newblock Recognition in terra incognita.
\newblock In \emph{ECCV}, 2018.

\bibitem[Bergmann et~al.(2019)Bergmann, Fauser, Sattlegger, and Steger]{bergmann2019mvtec}
Paul Bergmann, Michael Fauser, David Sattlegger, and Carsten Steger.
\newblock Mvtec ad -- a comprehensive real-world dataset for unsupervised anomaly detection.
\newblock In \emph{CVPR}, 2019.

\bibitem[Bolya et~al.(2026)Bolya, Huang, Sun, Cho, Madotto, Wei, Ma, Zhi, Rajasegaran, Bangalath, et~al.]{bolya2026perception}
Daniel Bolya, Po-Yao Huang, Peize Sun, Jang~Hyun Cho, Andrea Madotto, Chen Wei, Tengyu Ma, Jiale Zhi, Jathushan Rajasegaran, Hanoona Bangalath, et~al.
\newblock Perception encoder: The best visual embeddings are not at the output of the network.
\newblock \emph{NeurIPS}, 38:\penalty0 60884--60937, 2026.

\bibitem[Caron et~al.(2021)Caron, Touvron, Misra, J{\'e}gou, Mairal, Bojanowski, and Joulin]{caron2021emerging}
Mathilde Caron, Hugo Touvron, Ishan Misra, Herv{\'e} J{\'e}gou, Julien Mairal, Piotr Bojanowski, and Armand Joulin.
\newblock Emerging properties in self-supervised vision transformers.
\newblock In \emph{ICCV}, 2021.

\bibitem[Chen et~al.(2021)Chen, Liu, Zhao, and Jia]{chen2021distilling}
Pengguang Chen, Shu Liu, Hengshuang Zhao, and Jiaya Jia.
\newblock Distilling knowledge via knowledge review.
\newblock In \emph{CVPR}, 2021.

\bibitem[Cheng et~al.(2017)Cheng, Han, and Lu]{resisc45}
Gong Cheng, Junwei Han, and Xiaoqiang Lu.
\newblock Remote sensing image scene classification: Benchmark and state of the art.
\newblock \emph{Proceedings of the IEEE}, 105\penalty0 (10):\penalty0 1865--1883, Oct 2017.
\newblock ISSN 1558-2256.
\newblock \doi{10.1109/jproc.2017.2675998}.
\newblock URL \url{http://dx.doi.org/10.1109/JPROC.2017.2675998}.

\bibitem[Christie et~al.(2018)Christie, Fendley, Wilson, and Mukherjee]{fmow}
Gordon Christie, Neil Fendley, James Wilson, and Ryan Mukherjee.
\newblock Functional map of the world.
\newblock In \emph{CVPR}, 2018.

\bibitem[Chuang et~al.(2026)Chuang, Li, Wang, Yeh, Lyu, Raghavendra, Glass, Huang, Weston, Zettlemoyer, et~al.]{chuang2026meta}
Yung-Sung Chuang, Yang Li, Dong Wang, Ching-Feng Yeh, Kehan Lyu, Ramya Raghavendra, Jim Glass, Lifei Huang, Jason Weston, Luke Zettlemoyer, et~al.
\newblock Meta clip 2: A worldwide scaling recipe.
\newblock \emph{NeurIPS}, 38:\penalty0 48009--48036, 2026.

\bibitem[Deng et~al.(2009)Deng, Dong, Socher, Li, Li, and Fei-Fei]{deng2009imagenet}
Jia Deng, Wei Dong, Richard Socher, Li-Jia Li, Kai Li, and Li~Fei-Fei.
\newblock Imagenet: A large-scale hierarchical image database.
\newblock In \emph{CVPR}, pp.\  248--255, 2009.

\bibitem[Gao et~al.(2025)Gao, Wang, Tan, Xu, Liu, Hu, Chao, Zhang, and Li]{gao2025pfmbench}
Zhangyang Gao, Hao Wang, Cheng Tan, Chenrui Xu, Mengdi Liu, Bozhen Hu, Linlin Chao, Xiaoming Zhang, and Stan~Z Li.
\newblock Pfmbench: Protein foundation model benchmark.
\newblock \emph{arXiv preprint arXiv:2506.14796}, 2025.

\bibitem[Garrido et~al.(2023)Garrido, Balestriero, Najman, and Lecun]{garrido2023rankme}
Quentin Garrido, Randall Balestriero, Laurent Najman, and Yann Lecun.
\newblock Rankme: Assessing the downstream performance of pretrained self-supervised representations by their rank.
\newblock In \emph{ICML}. PMLR, 2023.

\bibitem[He \& Ozay(2022)He and Ozay]{he2022exploring}
Bobby He and Mete Ozay.
\newblock Exploring the gap between collapsed \& whitened features in self-supervised learning.
\newblock In \emph{ICML}. PMLR, 2022.

\bibitem[Helber et~al.(2019)Helber, Bischke, Dengel, and Borth]{helber2019eurosat}
Patrick Helber, Benjamin Bischke, Andreas Dengel, and Damian Borth.
\newblock Eurosat: A novel dataset and deep learning benchmark for land use and land cover classification.
\newblock \emph{IEEE Journal of Selected Topics in Applied Earth Observations and Remote Sensing}, 2019.

\bibitem[Heo et~al.(2019)Heo, Kim, Yun, Park, Kwak, and Choi]{heo2019comprehensive}
Byeongho Heo, Jeesoo Kim, Sangdoo Yun, Hyojin Park, Nojun Kwak, and Jin~Young Choi.
\newblock A comprehensive overhaul of feature distillation.
\newblock In \emph{ICCV}, pp.\  1921--1930, 2019.

\bibitem[Hinton et~al.(2015)Hinton, Vinyals, and Dean]{hinton2015distilling}
Geoffrey Hinton, Oriol Vinyals, and Jeff Dean.
\newblock Distilling the knowledge in a neural network.
\newblock \emph{arXiv preprint arXiv:1503.02531}, 2015.

\bibitem[Howard et~al.(2019)Howard, Sandler, Chu, Chen, Chen, Tan, Wang, Zhu, Pang, Vasudevan, et~al.]{howard2019searching}
Andrew Howard, Mark Sandler, Grace Chu, Liang-Chieh Chen, Bo~Chen, Mingxing Tan, Weijun Wang, Yukun Zhu, Ruoming Pang, Vijay Vasudevan, et~al.
\newblock Searching for mobilenetv3.
\newblock In \emph{ICCV}, 2019.

\bibitem[Jang et~al.(2025)Jang, Ma, and Lee]{jang2025vl2lite}
Jinseong Jang, Chunfei Ma, and Byeongwon Lee.
\newblock Vl2lite: Task-specific knowledge distillation from large vision-language models to lightweight networks.
\newblock In \emph{CVPR}, 2025.

\bibitem[Jing et~al.(2021)Jing, Vincent, LeCun, and Tian]{jing2021understanding}
Li~Jing, Pascal Vincent, Yann LeCun, and Yuandong Tian.
\newblock Understanding dimensional collapse in contrastive self-supervised learning.
\newblock \emph{arXiv preprint arXiv:2110.09348}, 2021.

\bibitem[Krizhevsky(2009)]{cifar}
Alex Krizhevsky.
\newblock Learning multiple layers of features from tiny images.
\newblock Technical report, University of Toronto, 2009.

\bibitem[Lee et~al.(2025)Lee, Das, Hayat, Choi, Hwang, and Porikli]{lee2025customkd}
Jungsoo Lee, Debasmit Das, Munawar Hayat, Sungha Choi, Kyuwoong Hwang, and Fatih Porikli.
\newblock Customkd: Customizing large vision foundation for edge model improvement via knowledge distillation.
\newblock In \emph{CVPR}, 2025.

\bibitem[Lee et~al.(2018)Lee, Kim, and Song]{lee2018self}
Seung~Hyun Lee, Dae~Ha Kim, and Byung~Cheol Song.
\newblock Self-supervised knowledge distillation using singular value decomposition.
\newblock In \emph{ECCV}, 2018.

\bibitem[Li et~al.(2022)Li, Yuan, Lu, Li, Chen, Engqvist, Kerkhoven, and Nielsen]{li2022deep}
Feiran Li, Le~Yuan, Hongzhong Lu, Gang Li, Yu~Chen, Martin~KM Engqvist, Eduard~J Kerkhoven, and Jens Nielsen.
\newblock Deep learning-based k cat prediction enables improved enzyme-constrained model reconstruction.
\newblock \emph{Nature catalysis}, 5\penalty0 (8):\penalty0 662--672, 2022.

\bibitem[Lin et~al.(2023)Lin, Akin, Rao, Hie, Zhu, Lu, Smetanin, Verkuil, Kabeli, Shmueli, dos Santos~Costa, Fazel-Zarandi, Sercu, Candido, and Rives]{lin2023evolutionary}
Zeming Lin, Halil Akin, Roshan Rao, Brian Hie, Zhongkai Zhu, Wenting Lu, Nikita Smetanin, Robert Verkuil, Ori Kabeli, Yaniv Shmueli, Allan dos Santos~Costa, Maryam Fazel-Zarandi, Tom Sercu, Salvatore Candido, and Alexander Rives.
\newblock Evolutionary-scale prediction of atomic-level protein structure with a language model.
\newblock \emph{Science}, 379\penalty0 (6637):\penalty0 1123--1130, 2023.
\newblock \doi{10.1126/science.ade2574}.

\bibitem[Lo~Conte et~al.(2000)Lo~Conte, Ailey, Hubbard, Brenner, Murzin, and Chothia]{lo2000scop}
Loredana Lo~Conte, Bart Ailey, Tim~JP Hubbard, Steven~E Brenner, Alexey~G Murzin, and Cyrus Chothia.
\newblock Scop: a structural classification of proteins database.
\newblock \emph{Nucleic acids research}, 28\penalty0 (1):\penalty0 257--259, 2000.

\bibitem[Loshchilov \& Hutter(2019)Loshchilov and Hutter]{loshchilov2019decoupled}
Ilya Loshchilov and Frank Hutter.
\newblock Decoupled weight decay regularization.
\newblock In \emph{ICLR}, 2019.

\bibitem[Maji et~al.(2013)Maji, Kannala, Rahtu, Blaschko, and Vedaldi]{maji2013fine}
Subhransu Maji, Juho Kannala, Esa Rahtu, Matthew Blaschko, and Andrea Vedaldi.
\newblock Fine-grained visual classification of aircraft.
\newblock \emph{arXiv preprint arXiv:1306.5151}, 2013.

\bibitem[Miles \& Mikolajczyk(2024)Miles and Mikolajczyk]{miles2024understanding}
Roy Miles and Krystian Mikolajczyk.
\newblock Understanding the role of the projector in knowledge distillation.
\newblock In \emph{AAAI}, 2024.

\bibitem[Miles et~al.(2024)Miles, Elezi, and Deng]{miles2024v}
Roy Miles, Ismail Elezi, and Jiankang Deng.
\newblock $ v\_ $\{$k$\}$ d $: Improving knowledge distillation using orthogonal projections.
\newblock In \emph{CVPR}. IEEE, 2024.

\bibitem[Oquab et~al.(2023)Oquab, Darcet, Moutakanni, Vo, Szafraniec, Khalidov, Fernandez, Haziza, Massa, El-Nouby, et~al.]{oquab2023dinov2}
Maxime Oquab, Timoth{\'e}e Darcet, Th{\'e}o Moutakanni, Huy Vo, Marc Szafraniec, Vasil Khalidov, Pierre Fernandez, Daniel Haziza, Francisco Massa, Alaaeldin El-Nouby, et~al.
\newblock Dinov2: Learning robust visual features without supervision.
\newblock \emph{arXiv preprint arXiv:2304.07193}, 2023.

\bibitem[Park et~al.(2019)Park, Kim, Lu, and Cho]{park2019relational}
Wonpyo Park, Dongju Kim, Yan Lu, and Minsu Cho.
\newblock Relational knowledge distillation.
\newblock In \emph{CVPR}, 2019.

\bibitem[Radford et~al.(2021)Radford, Kim, Hallacy, Ramesh, Goh, Agarwal, Sastry, Askell, Mishkin, Clark, Krueger, and Sutskever]{radford2021learning}
Alec Radford, Jong~Wook Kim, Chris Hallacy, Aditya Ramesh, Gabriel Goh, Sandhini Agarwal, Girish Sastry, Amanda Askell, Pamela Mishkin, Jack Clark, Gretchen Krueger, and Ilya Sutskever.
\newblock Learning transferable visual models from natural language supervision.
\newblock In \emph{ICML}, 2021.

\bibitem[Ranzinger et~al.(2024{\natexlab{a}})Ranzinger, Barker, Heinrich, Molchanov, Catanzaro, and Tao]{ranzinger2024phi}
Mike Ranzinger, Jon Barker, Greg Heinrich, Pavlo Molchanov, Bryan Catanzaro, and Andrew Tao.
\newblock Phi-s: Distribution balancing for label-free multi-teacher distillation.
\newblock \emph{arXiv preprint arXiv:2410.01680}, 2024{\natexlab{a}}.

\bibitem[Ranzinger et~al.(2024{\natexlab{b}})Ranzinger, Heinrich, Kautz, and Molchanov]{ranzinger2024radio}
Mike Ranzinger, Greg Heinrich, Jan Kautz, and Pavlo Molchanov.
\newblock Am-radio: Agglomerative vision foundation model reduce all domains into one.
\newblock In \emph{CVPR}, 2024{\natexlab{b}}.

\bibitem[Romero et~al.(2015)Romero, Ballas, Ebrahimi~Kahou, Chassang, Gatta, and Bengio]{romero2015fitnets}
Adriana Romero, Nicolas Ballas, Samira Ebrahimi~Kahou, Antoine Chassang, Carlo Gatta, and Yoshua Bengio.
\newblock {FitNets}: Hints for thin deep nets, 2015.
\newblock URL \url{https://arxiv.org/abs/1412.6550}.

\bibitem[Sar{\i}y{\i}ld{\i}z et~al.(2024)Sar{\i}y{\i}ld{\i}z, Weinzaepfel, Lucas, Larlus, and Kalantidis]{sariyildiz2024unic}
Mert~B{\"u}lent Sar{\i}y{\i}ld{\i}z, Philippe Weinzaepfel, Thomas Lucas, Diane Larlus, and Yannis Kalantidis.
\newblock Unic: Universal classification models via multi-teacher distillation.
\newblock In \emph{ECCV}. Springer, 2024.

\bibitem[Sim{\'e}oni et~al.(2025)Sim{\'e}oni, Vo, Seitzer, Baldassarre, Oquab, Jose, Khalidov, Szafraniec, Yi, Ramamonjisoa, et~al.]{simeoni2025dinov3}
Oriane Sim{\'e}oni, Huy~V Vo, Maximilian Seitzer, Federico Baldassarre, Maxime Oquab, Cijo Jose, Vasil Khalidov, Marc Szafraniec, Seungeun Yi, Micha{\"e}l Ramamonjisoa, et~al.
\newblock Dinov3.
\newblock \emph{arXiv preprint arXiv:2508.10104}, 2025.

\bibitem[Thumuluri et~al.(2022)Thumuluri, Almagro~Armenteros, Johansen, Nielsen, and Winther]{thumuluri2022deeploc}
Vineet Thumuluri, Jos{\'e}~Juan Almagro~Armenteros, Alexander~Rosenberg Johansen, Henrik Nielsen, and Ole Winther.
\newblock Deeploc 2.0: multi-label subcellular localization prediction using protein language models.
\newblock \emph{Nucleic acids research}, 50\penalty0 (W1):\penalty0 W228--W234, 2022.

\bibitem[Tian et~al.(2020)Tian, Krishnan, and Isola]{tian2020contrastive}
Yonglong Tian, Dilip Krishnan, and Phillip Isola.
\newblock Contrastive representation distillation.
\newblock In \emph{ICLR}, 2020.

\bibitem[Van~Horn et~al.(2018)Van~Horn, Mac~Aodha, Song, Cui, Sun, Shepard, Adam, Perona, and Belongie]{van2018inaturalist}
Grant Van~Horn, Oisin Mac~Aodha, Yang Song, Yin Cui, Chen Sun, Alex Shepard, Hartwig Adam, Pietro Perona, and Serge Belongie.
\newblock The inaturalist species classification and detection dataset.
\newblock In \emph{CVPR}, 2018.

\bibitem[Vemulapalli et~al.(2024)Vemulapalli, Pouransari, Faghri, Mehta, Farajtabar, Rastegari, and Tuzel]{vemulapalli2024knowledge}
Raviteja Vemulapalli, Hadi Pouransari, Fartash Faghri, Sachin Mehta, Mehrdad Farajtabar, Mohammad Rastegari, and Oncel Tuzel.
\newblock Knowledge transfer from vision foundation models for efficient training of small task-specific models.
\newblock In \emph{ICML}, 2024.

\bibitem[Wah et~al.(2011)Wah, Branson, Welinder, Perona, and Belongie]{cub}
Catherine Wah, Steve Branson, Peter Welinder, Pietro Perona, and Serge Belongie.
\newblock The caltech-ucsd birds-200-2011 dataset.
\newblock \emph{California Institute of Technology}, 2011.

\bibitem[Wang et~al.(2019)Wang, Yuan, Zhang, and Feng]{wang2019distilling}
Tao Wang, Li~Yuan, Xiaopeng Zhang, and Jiashi Feng.
\newblock Distilling object detectors with fine-grained feature imitation.
\newblock In \emph{CVPR}, 2019.

\bibitem[Xu et~al.(2024)Xu, Xie, Tan, Huang, Howes, Sharma, Li, Ghosh, Zettlemoyer, and Feichtenhofer]{xu2024demystifying}
Hu~Xu, Saining Xie, Xiaoqing Tan, Po-Yao Huang, Russell Howes, Vasu Sharma, Shang-Wen Li, Gargi Ghosh, Luke Zettlemoyer, and Christoph Feichtenhofer.
\newblock Demystifying clip data.
\newblock In \emph{ICLR}, volume 2024, pp.\  47812--47831, 2024.

\bibitem[Yang et~al.(2023)Yang, Shi, Wei, Liu, Zhao, Ke, Pfister, and Ni]{medmnistv2}
Jiancheng Yang, Rui Shi, Donglai Wei, Zequan Liu, Lin Zhao, Bilian Ke, Hanspeter Pfister, and Bingbing Ni.
\newblock Medmnist v2-a large-scale lightweight benchmark for 2d and 3d biomedical image classification.
\newblock \emph{Scientific Data}, 10\penalty0 (1):\penalty0 41, 2023.

\bibitem[Zbontar et~al.(2021)Zbontar, Jing, Misra, LeCun, and Deny]{zbontar2021barlow}
Jure Zbontar, Li~Jing, Ishan Misra, Yann LeCun, and St{\'e}phane Deny.
\newblock Barlow twins: Self-supervised learning via redundancy reduction.
\newblock In \emph{ICML}. PMLR, 2021.

\bibitem[Zhang et~al.(2025)Zhang, Ma, Bai, Wang, and Fu]{zhang2025accessing}
Yitian Zhang, Xu~Ma, Yue Bai, Huan Wang, and Yun Fu.
\newblock Accessing vision foundation models via imagenet-1k.
\newblock In \emph{ICLR}, 2025.

\bibitem[Zou et~al.(2022)Zou, Jeong, Pemula, Zhang, and Dabeer]{zou2022visa}
Yang Zou, Jongheon Jeong, Latha Pemula, Dongqing Zhang, and Onkar Dabeer.
\newblock Spot-the-difference self-supervised pre-training for anomaly detection and segmentation.
\newblock In \emph{ECCV}, 2022.

\end{thebibliography}
\bibliographystyle{iclr2027_conference}

\appendix
\section{Appendix}
\section{Theoretical Motivation}
\label{sec:theory}

\noindent\textbf{Insights into vanilla feature matching. }
We analyze conventional feature matching from the perspective of the
principal components of the teacher representation.
Let $z_b^t \in \mathbb{R}^d$ denote the centered teacher
representation for the $b$-th sample, and let
$\tilde{z}_b^s \in \mathbb{R}^d$ denote the corresponding student
prediction.
We apply PCA to the teacher representations and denote the resulting
orthonormal principal directions by
$\{u_k\}_{k=1}^{d}$.
The teacher representation can then be written as
\begin{equation}
    z_b^t
    =
    \sum_{k=1}^{d} c_{b,k}^t u_k,
\end{equation}
where
\begin{equation}
    c_{b,k}^t = u_k^\top z_b^t.
\end{equation}
We similarly denote the student prediction along the $k$-th principal
direction by $\tilde{c}_{b,k}^s$.
Since the PCA basis is orthonormal, projecting both representations onto
this basis does not change the conventional feature matching objective:
\begin{equation}
\begin{aligned}
    \left\|
        \tilde{\bm{z}}_b^s - \bm{z}_b^t
    \right\|_2^2
    &=
    \sum_{k=1}^{d}
    \left(
        \tilde{c}_{b,k}^s - c_{b,k}^t
    \right)^2.
\end{aligned}
\end{equation}
Thus, expressing feature matching in the PCA basis is equivalent to
conventional feature matching.

The standard feature matching objective is given by
\begin{equation}
    \mathcal{L}_{\mathrm{FM}}
    =\sum_{k=1}^{d} e_k,
    \qquad
    e_k
    =
    \frac{1}{B}\sum_{b=1}^{B}
    \left(
        \tilde{c}_{b,k}^s-c_{b,k}^t
    \right)^2,
\end{equation}
where $e_k$ denotes the batch-averaged reconstruction error along the
$k$-th principal direction.

\paragraph{Reconstruction error and teacher variance.}
Let
\begin{equation}
    \lambda_k = \operatorname{Var}(c_k^t),
    \qquad
    v_k = \operatorname{Var}(\tilde{c}_k^s),
\end{equation}
and let $\rho_k$ denote the correlation coefficient between
$c_k^t$ and $\tilde{c}_k^s$.
Furthermore, define the difference between their means as
\begin{equation}
    b_k
    =
    \mathbb{E}[\tilde{c}_k^s]
    -
    \mathbb{E}[c_k^t].
\end{equation}
The expected reconstruction error satisfies
\begin{equation}
\begin{aligned}
    \mathbb{E}[e_k]
    &=
    \mathbb{E}
    \left[
        \left(
            \tilde{c}_k^s-c_k^t
        \right)^2
    \right] \\
    &=
    \lambda_k
    +
    v_k
    -
    2\rho_k\sqrt{\lambda_k v_k}
    +
    b_k^2.
    \label{eq:expected_error}
\end{aligned}
\end{equation}

At the early stage of training, the student prediction is largely
uninformative about the teacher principal directions.
It is therefore reasonable to assume that
\begin{equation}
    \rho_k \approx 0,
    \qquad
    v_k \approx v,
    \qquad
    b_k \approx b,
\end{equation}
where $v$ and $b$ are approximately constant across principal
components.
Under these assumptions,
\begin{equation}
    \mathbb{E}[e_k]
    \approx
    \lambda_k + C,
    \qquad
    C = v+b^2.
   \label{eq:error_lambda_app}
\end{equation}
Therefore, at the early stage of training, the reconstruction error is
approximately an affine function of the teacher variance.

\paragraph{Dominance in optimization.}
Let
\begin{equation}
    r_{b,k}
    =
    \tilde{c}_{b,k}^s-c_{b,k}^t
\end{equation}
denote the residual along the $k$-th principal direction.
The gradient of the feature matching objective with respect to the
student prediction is
\begin{equation}
    \frac{\partial \mathcal{L}_{\mathrm{FM}}}
         {\partial \tilde{c}_{b,k}^s}
    =
    \frac{2}{Bd}r_{b,k}.
\end{equation}
Consequently, the squared norm of the output-space gradient associated
with the $k$-th principal direction is
\begin{equation}
\begin{aligned}
    \left\|
        \nabla_{\tilde{\bm{c}}_{:,k}^s}
        \mathcal{L}_{\mathrm{FM}}
    \right\|_2^2
    &=
    \sum_{b=1}^{B}
    \left(
        \frac{2}{Bd}r_{b,k}
    \right)^2 \\
    &=
    \frac{4}{B d^2} e_k.
    \label{eq:gradient_error}
\end{aligned}
\end{equation}
Thus, the gradient magnitude contributed by each principal direction is
directly proportional to its reconstruction error.
Combining Equation~\ref{eq:error_lambda_app} and~\ref{eq:gradient_error}
gives
\begin{equation}
    \mathbb{E}
    \left[
        \left\|
            \nabla_{\tilde{\bm{c}}_{:,k}^s}
            \mathcal{L}_{\mathrm{FM}}
        \right\|_2^2
    \right]
    \approx
    \frac{4}{B d^2}
    \left(
        \lambda_k+C
    \right).
\end{equation}
Hence, high-variance principal components produce larger reconstruction
errors and stronger optimization signals, causing conventional feature
matching to preferentially fit the dominant spectral directions of the teacher representation.

\begin{figure}[t]
    \centering
       \includegraphics[height=0.3\linewidth]{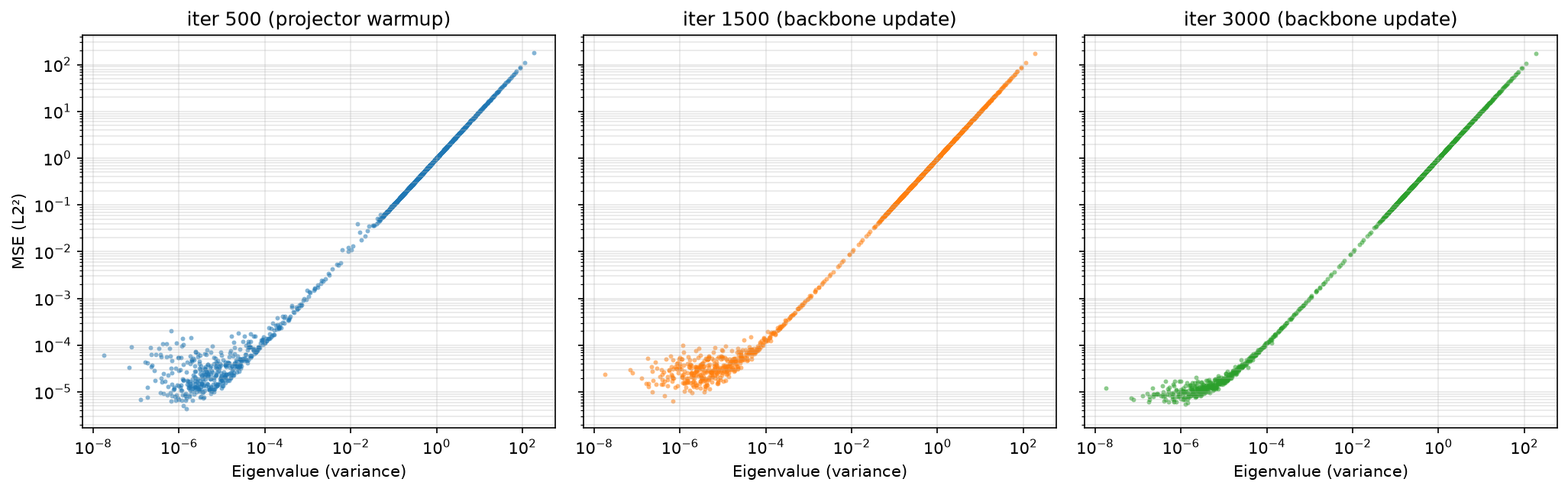}
    \caption{\small Eigenvalues (x-axis) and reconstruction error.}
    \label{fig:eigenvalue_mse}
\end{figure}
\begin{figure}[t]
    \centering
       \includegraphics[height=0.3\linewidth]{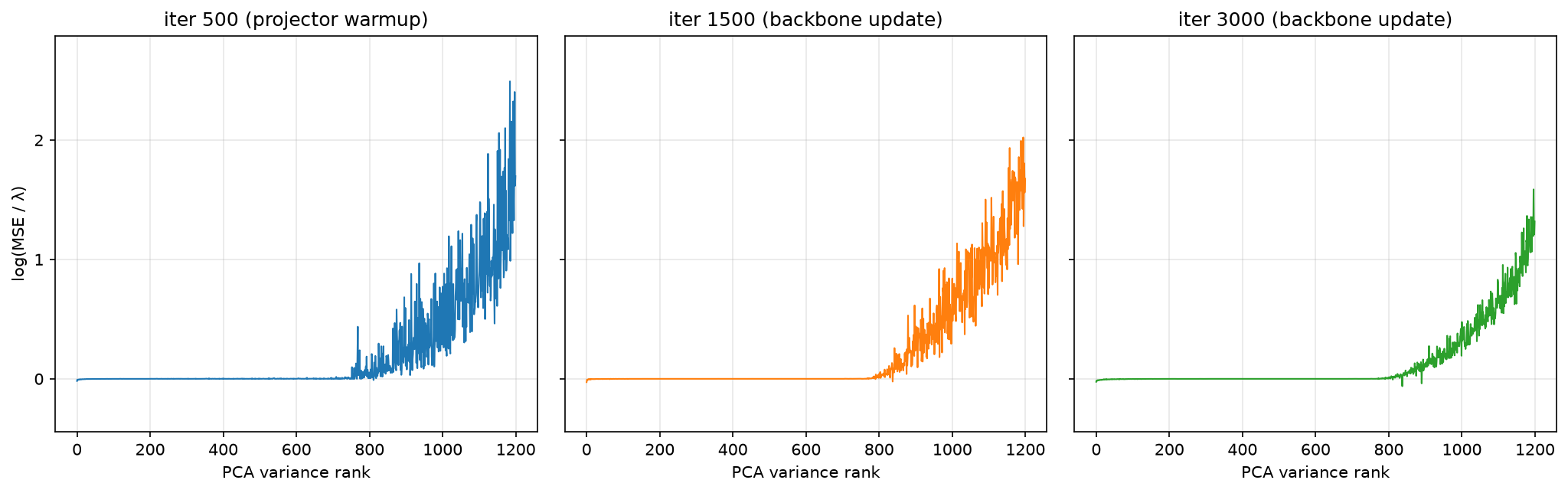}
    \caption{\small Reconstruction error along principal components.}
    \label{fig:mse_iterations}
\end{figure}

\paragraph{Empirical analysis.}
Fig.~\ref{fig:eigenvalue_mse} describes the eigenvalues and corresponding error in the axis. Fig.~\ref{fig:mse_iterations} describes the reconstruction error along principal components with L2-distance feature matching loss. Fig.~\ref{fig:eigenvalue_mse} empirically supports the Equation~\ref{eq:error_lambda_app}, i.e., the error is proportional to the eigenvalues. Also, the trend is consistent across different training iterations.

\begin{algorithm}[t]
\caption{PyTorch-style pseudocode for \ours.}
\label{alg:specmatch}
\begin{lstlisting}[language=Python,basicstyle=\ttfamily\small]
def specmatch_loss(student, teacher, beta_inv=3.0, eps=1e-3):
    # student and teacher: PCA coefficients shaped (B, D)."""
    # reconstruction error for each principal direction
    error = (student - teacher).pow(2).mean(dim=0)
    beta = 1.0 / beta_inv
    # mitigate the imbalance across principal directions
    loss = (error + eps).pow(beta).mean()
    return loss
\end{lstlisting}
\end{algorithm}
\section{Experimetal Details}
\label{sec:exp_details}

We provide additional details on the datasets and experimental settings used in our experiments. Table~\ref{tab:dataset_statistics} summarizes the datasets used in the experiments reported in Table~1. We report the number of training and test samples and the number of classes for each dataset. 
Similarly, Table~\ref{tab:protein_statistics} summarizes the statistics of the protein datasets used in the experiments reported in Table~\ref{tab:protein_results}.

\begin{table}[t]
    \centering
    \caption{Dataset statistics.}
    \label{tab:dataset_statistics}
    \begin{tabular}{lrrr}
        \toprule
        \textbf{Dataset} & \textbf{Train} & \textbf{Test} & \textbf{\# Classes} \\
        \midrule
        CIFAR-100   & 50,000  & 10,000 & 100 \\
        CIFAR-10    & 50,000  & 10,000 & 10  \\
        SVHN        & 73,257  & 26,032 & 10  \\
        GTSRB       & 26,640  & 12,630 & 43  \\
        RESISC45    & 18,900  & 6,300  & 45  \\
        EuroSAT     & 21,600  & 5,400  & 10  \\
        OrganCMNIST & 12,975  & 8,216  & 11  \\
        Camelyon17  & 302,436 & 85,054 & 2   \\
        VisA        & 10,821   & 962 normal + 1,200 anomaly  & 2   \\
        MVTec       & 5,354   & 467 normal + 1,258 anomaly  & 2   \\
        iWildCam    & 129,809 & 42,791 & 182 \\
        FMoW        & 76,863  & 22,108 & 62  \\
        CUB         & 5,994   & 5,794  & 200 \\
        NABirds     & 23,929  & 24,633 & 555 \\
        \bottomrule
    \end{tabular}
\end{table}

\begin{table}[t]
\centering
\caption{Statistics of the protein datasets used in Table~\ref{tab:protein_results}.}
\label{tab:protein_statistics}
\resizebox{0.9\linewidth}{!}{
\begin{tabular}{lcccccc}
\toprule
Dataset & Task & Classes & Train & Val & Test & Epochs\\
\midrule
Fold prediction & classification &1195 & 13034&1628 & 1630 &20\\
GO molecular function & multi labels classification & 489 & 22291&2785&2787&20 \\
GO cellular component & multi labels classification & 320 & 11196&1398&1400&20 \\
Enzyme commission number &multi labels classification & 585&12928&1615&1616&60\\
Enzyme catalytic efficiency & regression&-&10363&1290&1298&60\\
DeepLoc2Multi & multi labels classification&10&21949&2743&2744&50\\
\bottomrule
\end{tabular}}
\end{table}

\noindent\textbf{Experimental settings for Table~\ref{tab:main_results}.} We use AdamW as the optimizer in all experiments. The learning rate for the backbone is set to $1\times10^{-4}$ by default, while we use a smaller learning rate of $1\times10^{-5}$ for MVTec and iWildCam. The learning rate for the projector is set to $1\times10^{-3}$. We first trained only the linear projector for 500 iterations as a warm-up, and then jointly trained the backbone and projector for an additional 2,500 iterations. These hyperparameters are kept fixed across all three teacher--student combinations to ensure a consistent comparison. We set $\epsilon$ as 1e-3 and 1e-4 for the training on PE-Core-Tiny and MobileNet-V3, respectively. 

\noindent\textbf{Experiments on ViSA and MvTec.} During the distillation training using unlabeled data, we employ both normal and anomaly samples as training data. For the experiments on ViSA, we employ a few labeled normal and anomaly data to train a linear classifier following their evaluation protocol. For the experiments on MvTec, we employ the nearest neighbor distance to the normal samples to compute the anomaly score. 

\noindent\textbf{Experiments on ImageNet.} We use a batch size of 512 and train for 200,000 iterations. When initializing the model with PE-Core-Tiny, we use a fixed learning rate of $1\times10^{-5}$ throughout training. For training from scratch, we use an initial learning rate of $1\times10^{-3}$ and decay the learning rate using a cosine schedule.

\noindent\textbf{Protein Experiments.} For the protein-related experiments in Table~\ref{tab:protein_results}, we use AdamW with a learning rate of $1\times10^{-4}$. The learning rate is kept fixed throughout training. Since the number of training samples varies substantially across the protein datasets, we adjust the number of training epochs for each dataset accordingly. The dataset-specific training configurations are summarized in Table~\ref{tab:protein_statistics}.

\begin{table*}[t]
\centering
\caption{
Overview of teacher and student vision models used in our experiments.
Parameter counts correspond to the vision backbone only, excluding classification
heads and text encoders.
}
\label{tab:models}
\small
\setlength{\tabcolsep}{4pt}

\begin{tabular}{llllcc}
\toprule
Role & Model & Source & Pre-training & Input & Params. \\
\midrule

\multirow{3}{*}{Teacher}
& DINOv2 ViT-g/14
& \href{https://github.com/facebookresearch/dinov2}{DINOv2}
& DINOv2
& 518 & 1.1B \\

& PE-Core-G14-448
& \href{https://huggingface.co/facebook/PE-Core-G14-448}{Meta PE}
& Perception Encoder
& 448 & 1.9B \\


& CLIP ConvNeXt-XXLarge
& \href{https://huggingface.co/laion/CLIP-convnext_xxlarge-laion2B-s34B-b82K-augreg}{OpenCLIP}
& LAION-2B
& 256 & 846.5M \\

\midrule

\multirow{4}{*}{Student}
& MobileNetV3-Small
& \href{https://github.com/rwightman/pytorch-image-models/releases/download/v0.1-weights/mobilenetv3_small_100_lamb-266a294c.pth}{timm}
& IN-1K
& 224 & 1.5M \\


& PE-Core-T16-384
& \href{https://huggingface.co/facebook/PE-Core-T16-384}{Meta PE}
& Perception Encoder
& 384 & 10M \\

& ConvNeXt-Tiny
& \href{https://github.com/huggingface/pytorch-image-models}{timm}
& Scratch
& 224 & 27.8M \\
& ViT-Tiny/16
& \href{https://huggingface.co/timm/vit_tiny_patch16_224.augreg_in21k_ft_in1k}{timm}
& Scratch
& 224 & 5.5M \\

\bottomrule
\end{tabular}

\vspace{1mm}
\footnotesize
$^\dagger$ViT-Tiny/16 is also trained from scratch in experiments where specified.
Links in the Source column point to the model repository or the exact checkpoint used.
\end{table*}
\section{Additional Results}
\label{sec:additional_results}
\begin{figure*}[t]
    \centering
    \begin{subfigure}[t]{0.24\textwidth}
        \centering
        \includegraphics[width=\linewidth]{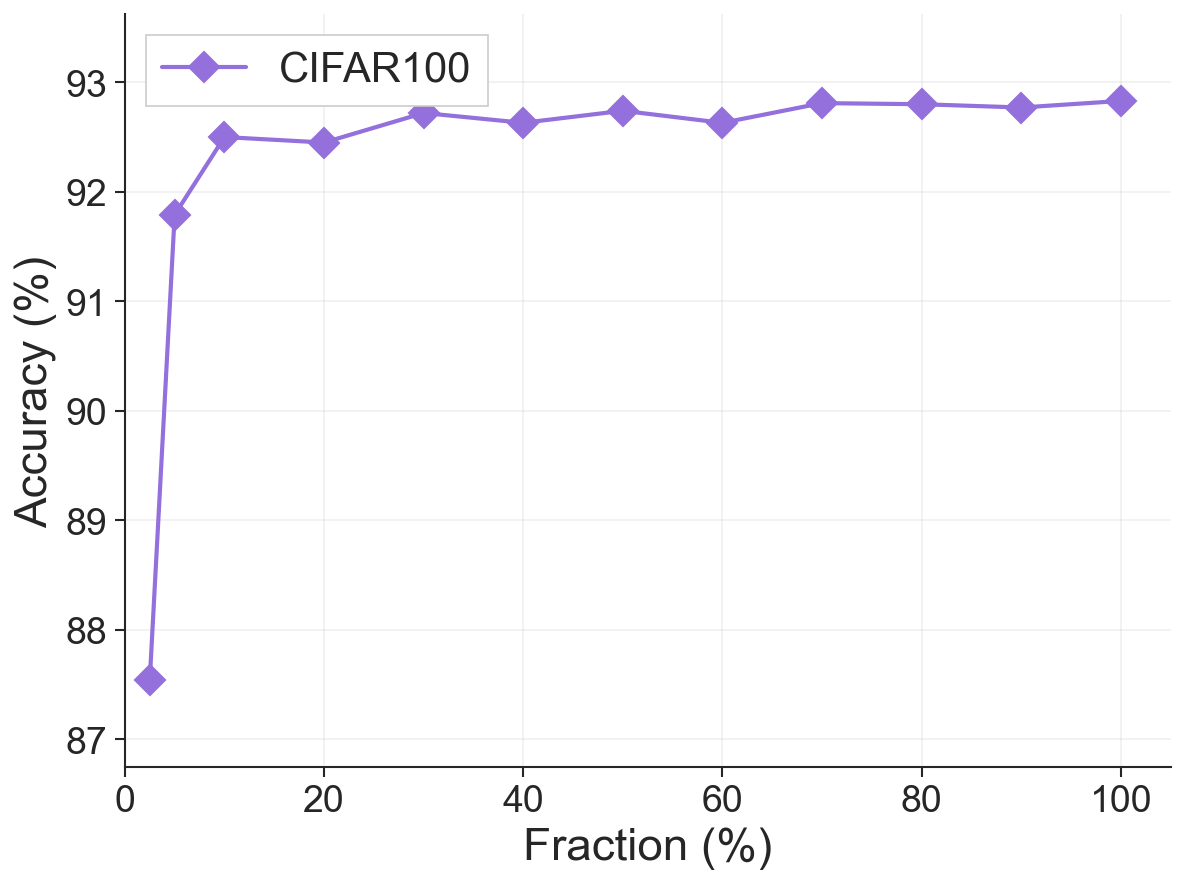}
        \caption{CIFAR100}
        \label{fig:graph1}
    \end{subfigure}
    \hfill
    \begin{subfigure}[t]{0.24\textwidth}
        \centering
        \includegraphics[width=\linewidth]{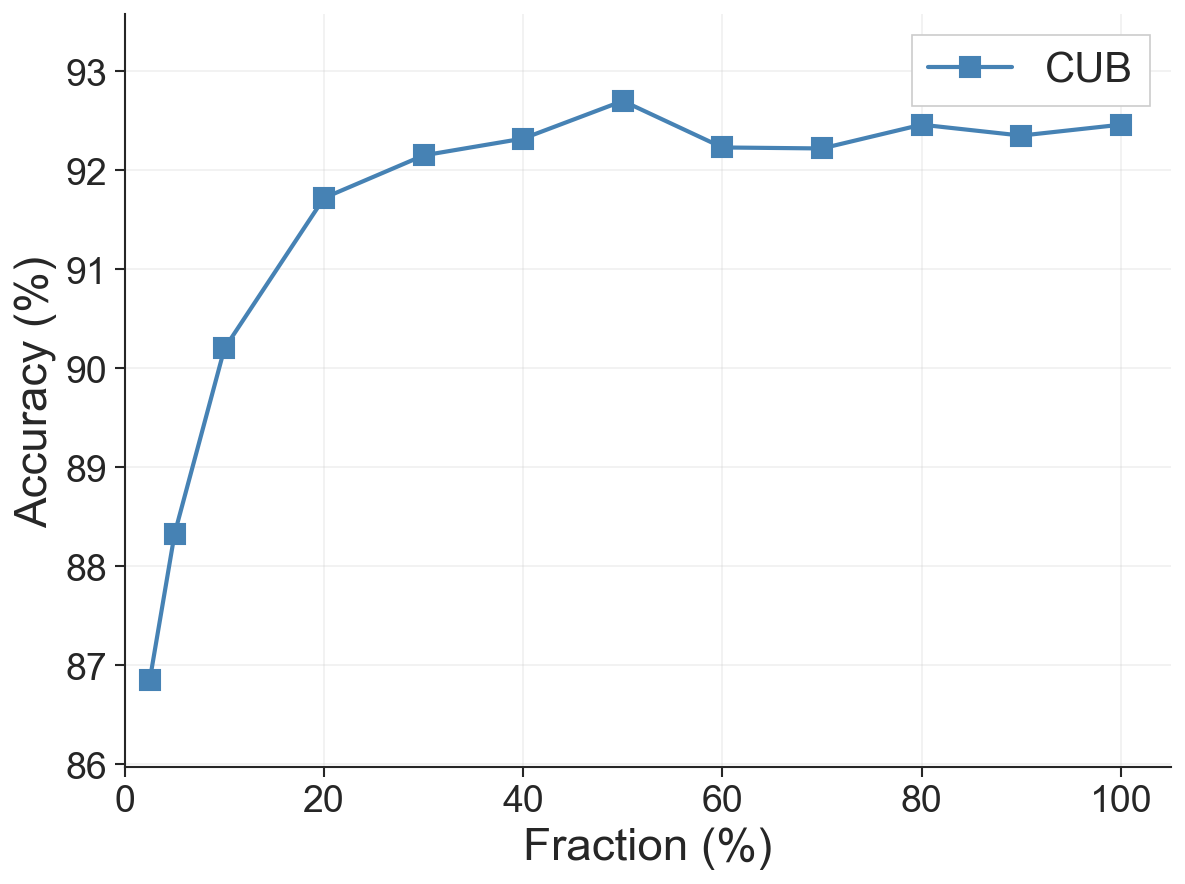}
        \caption{CUB}
        \label{fig:graph2}
    \end{subfigure}
    \hfill
    \begin{subfigure}[t]{0.24\textwidth}
        \centering
        \includegraphics[width=\linewidth]{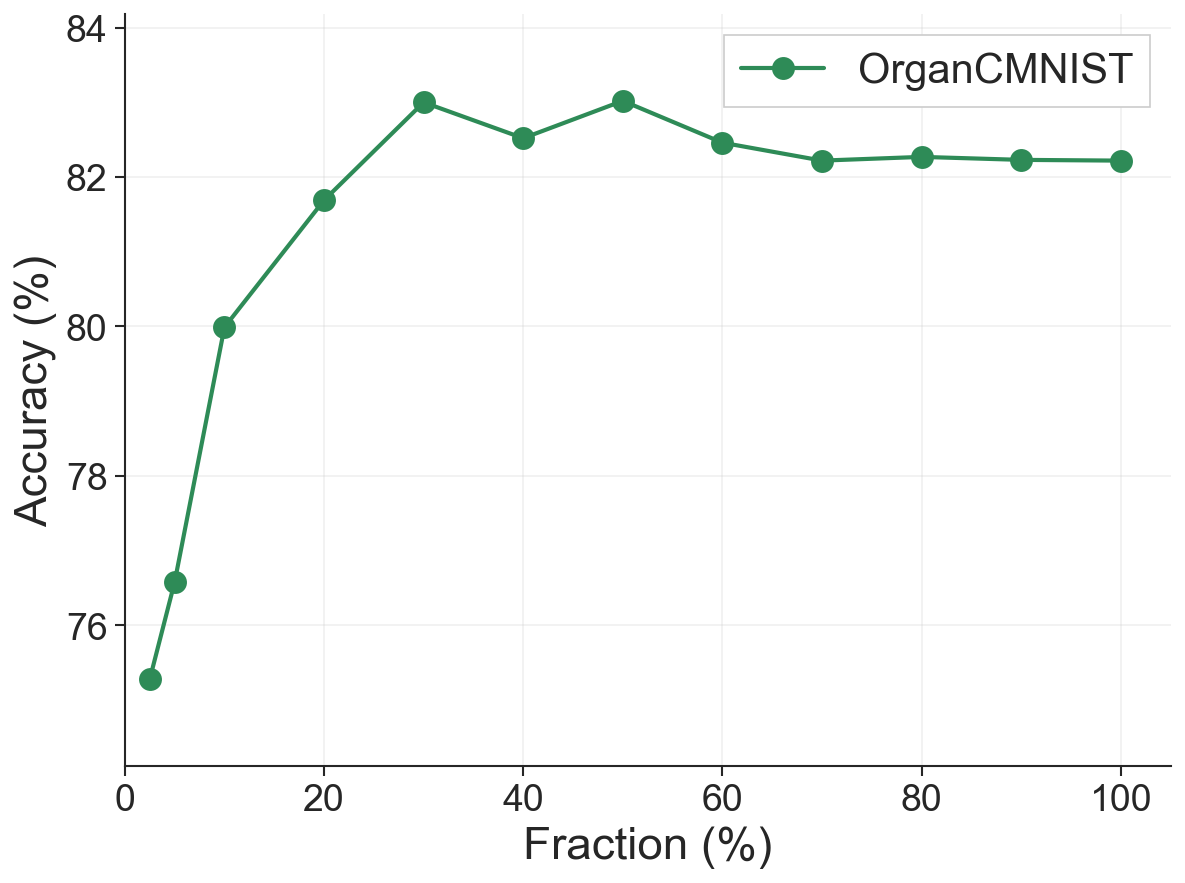}
        \caption{OrganCMNIST}
        \label{fig:graph3}
    \end{subfigure}
    \hfill
    \begin{subfigure}[t]{0.24\textwidth}
        \centering
        \includegraphics[width=\linewidth]{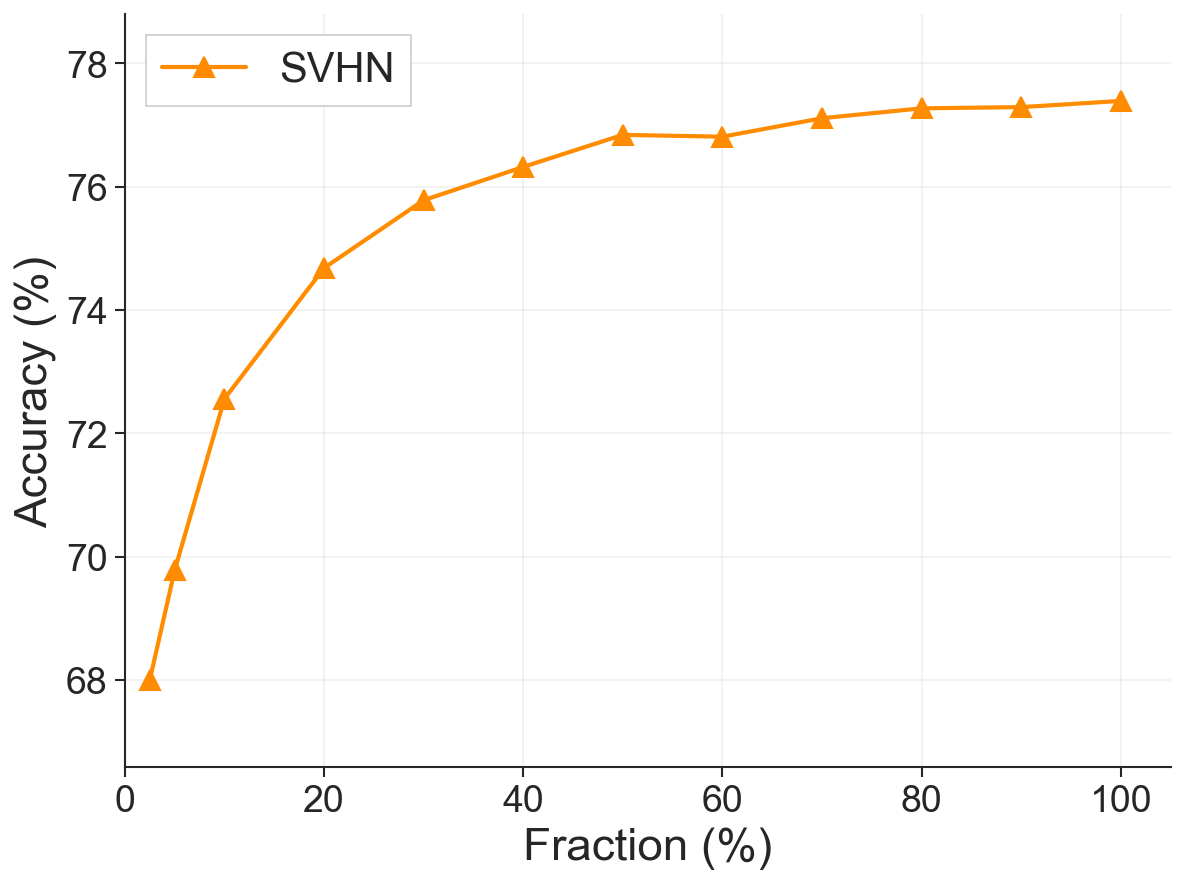}
        \caption{SVHN}
        \label{fig:graph4}
    \end{subfigure}
    \caption{Relationship between linear probe accuracy and the number of used principal components. We measure the linear probe accuracy in using top K\% principal components.}
    \label{fig:pca_percent_accuracy}
\end{figure*}
\noindent\textbf{The number of principal components and linear probe accuracy.} 
In relation to Figure 1, Fig.~\ref{fig:pca_percent_accuracy} studies how linear-probing accuracy changes as the number of principal components increases. For tasks such as CIFAR-100, high performance was achieved using only a small number of components. For many other tasks, however, performance continued to improve as up to approximately 50\% of the components were included. On SVHN, performance improved further as even more components were used. These results indicate that low-variance principal components can contain information useful for certain downstream tasks, rather than merely representing noise.

\noindent\textbf{Comparison between dynamic and static loss weighting.} In Sec~\ref{sec:method} (Eq.~\ref{eq:static_loss}), we mention the alternative formulation to realize the spectral balancing, which aims to mitigate the spectral bias by using the variance in the target representations. Table~\ref{tab:static_dynamic} compares the results by setting $\beta^{-1} = 3$ in both loss. Overall, the dynamic weighting scheme used in the main paper achieves better performance. On ImageNet, however, static weighting performs slightly better, suggesting that the difference between the two schemes is modest. Importantly, both weighting schemes consistently outperform the unweighted L2 loss, demonstrating that variance-based loss weighting improves performance.

\begin{table*}[t]
\centering
\small
\setlength{\tabcolsep}{4.0pt}
\caption{
Comparison with fully supervised fine-tuning.
All methods use the same pretrained MobileNetV3-Small student.
$\Delta$ denotes the performance difference between \ours and supervised fine-tuning.
}
\label{tab:supervised_finetuning}
\resizebox{\textwidth}{!}{
\begin{tabular}{lcccccccccccc}
\toprule
Method
& CIFAR100 & CIFAR10 & SVHN & GTSRB
& NABirds & CUB
& RESISC & EuroSAT
& OrganCMNIST & Camelyon17
& iWildCam & FMoW \\
\midrule

Supervised FT
& 76.5 & 93.8 & \textbf{95.2} & \textbf{95.4}
& 50.1 & 55.8
& 93.0 & \textbf{98.4}
& \textbf{93.4} & 90.1
& 58.5 & 41.0 \\

\midrule

\ours ($T$: PE-Core)
& 77.6 & 95.2 & 89.5 & 94.6
& \textbf{64.2} & \textbf{72.5}
& \textbf{93.6} & 98.2
& 90.7 & \textbf{93.9}
& \textbf{62.4} & \textbf{42.1} \\

$\Delta$
& +1.1 & +1.5 & -5.7 & -0.8
& +14.1 & +16.7
& +0.6 & -0.2
& -2.7 & +3.8
& +3.9 & +1.1 \\

\midrule

\ours ($T$: DINOv2)
& 78.1 & 95.5 & 82.3 & 88.1
& 56.8 & 69.0
& 92.8 & 98.0
& 91.0 & 93.4
& 61.1 & 39.5 \\

$\Delta$
& +1.6 & +1.8 & -12.9 & -7.3
& +6.7 & +13.2
& -0.2 & -0.4
& -2.4 & +3.3
& +2.6 & -1.5 \\

\bottomrule
\end{tabular}
}
\end{table*}

\begin{table*}[t]
\centering
\small
\caption{\small Instance recognition results on the Oxford and Paris datasets. A ConvNeXt-Large model pre-trained with image-text data is used as the teacher, while ConvNeXt-Tiny serves as the student.}
\vspace{-3mm}
\setlength{\tabcolsep}{3pt}
\resizebox{\textwidth}{!}{
\begin{tabular}{l|ccc|ccc|ccc|ccc|ccc|ccc}
\toprule
\multirow{3}{*}{Method}
& \multicolumn{9}{c|}{Oxford}
& \multicolumn{9}{c}{Paris} \\
\cmidrule(lr){2-10}\cmidrule(lr){11-19}
& \multicolumn{3}{c|}{Easy}
& \multicolumn{3}{c|}{Medium}
& \multicolumn{3}{c|}{Hard}
& \multicolumn{3}{c|}{Easy}
& \multicolumn{3}{c|}{Medium}
& \multicolumn{3}{c}{Hard} \\
\cmidrule(lr){2-4}\cmidrule(lr){5-7}\cmidrule(lr){8-10}
\cmidrule(lr){11-13}\cmidrule(lr){14-16}\cmidrule(lr){17-19}
& mAP & P@1 & P@10
& mAP & P@1 & P@10
& mAP & P@1 & P@10
& mAP & P@1 & P@10
& mAP & P@1 & P@10
& mAP & P@1 & P@10 \\\hline
\rowcolor{blue!10} ConvNEXT-XXlarge $\rightarrow$ ConvNEXT-Tiny  &&&&&&&&&&&&&&&&&&\\
DINO & \textbf{21.9}&\textbf{41.2}&\textbf{27.7} & 15.0&\textbf{45.7}&\textbf{27.3}&3.4&17.1&5.3&\textbf{64.8}&\textbf{95.7}&\textbf{87.7}&\textbf{49.9}&\textbf{97.1}&\textbf{91.3}&\textbf{24.7}&\textbf{68.6}&\textbf{56.0}\\
FM
&17.6&29.4&20.9
&13.8&30.0&18.7
&3.7&10.0&7.0
&53.4&87.1&81.0
&40.5&88.6&82.7
&17.4&54.3&41.6\\

\ours
&19.7&30.9&22.4
&\textbf{15.7}&38.6&22.3
&\textbf{4.9}&\textbf{18.6}&\textbf{8.6}
&55.5&87.1&81.6
&43.7&88.6&84.4
&21.5&67.1&49.9\\\hline

\rowcolor{blue!10} PE-Core-G14 $\rightarrow$ ViT-Tiny  &&&&&&&&&&&&&&&&&&\\
DINO&\textbf{19.0}&\textbf{36.7}&\textbf{28.1}&\textbf{14.4}&\textbf{37.1}&\textbf{26.9}&\textbf{2.9}&\textbf{8.6}&\textbf{4.0}&\textbf{60.3}&\textbf{95.7}&\textbf{88.3}&\textbf{45.2}&\textbf{98.6}&\textbf{90.4}&\textbf{19.8}&\textbf{68.6}&\textbf{51.4}\\
FM &10.0&14.7& 14.9&9.1&18.6&16.4&2.2&5.7&3.1&30.3&75.7&64.7&24.7&77.1&66.9&9.1&31.4&19.4\\

\ours &11.2&29.4&14.3& 9.6&31.4&15.3&2.4&5.7&2.7&32.9&77.1&64.9&26.8&78.6&67.6&11.2&35.7&23.4\\
\bottomrule
\end{tabular}}

\label{tab:instance_recognition}
\end{table*}

\noindent\textbf{Instance recognition.} To further evaluate the quality of the representations learned on ImageNet, we assess the ConvNeXt-Tiny model trained in the ImageNet experiments on the Oxford and Paris instance recognition benchmarks. Since this task requires discriminative instance-level representations rather than category-level classification, it provides a complementary evaluation of representation quality. As shown in Table~\ref{tab:instance_recognition}, \ours consistently outperforms vanilla feature matching across most metrics, indicating that it better preserves the teacher's rich feature representations. DINO outperforms \ours on many metrics, in contrast to the results on ImageNet (Table~\ref{tab:imagenet_shots}). This suggests that DINO is more effective at capturing fine-grained, instance-level details, whereas \ours is better suited to transferring class-discriminative features.

\begin{table}[t]
    \centering
\small
    \setlength{\tabcolsep}{3pt}
    \caption{Comparison between static and dynamic weighting.}
    \begin{tabular}{lcccccc}
        \toprule
        Method & CIFAR-10 & SVHN & NABirds & CUB & RESISC & ImageNet \\
        \midrule
        FM & 94.8&87.6&62.6&71.3&92.6&72.1\\
        Static  & 95.0 & 88.3 & 63.7 & 71.7 & 93.6 & \textbf{73.2} \\
        Dynamic & \textbf{95.2} & \textbf{89.5} & \textbf{64.2} &
                  \textbf{72.5} & 93.6 & 73.1 \\
        \bottomrule
    \end{tabular}
    \label{tab:static_dynamic}
\end{table}

\end{document}